\documentclass{article}

\usepackage[accepted]{icml2025}

\usepackage{microtype}
\usepackage{graphicx}
\usepackage{subfigure}
\usepackage{booktabs} 
\usepackage{hyperref}
\makeatletter
\@ifundefined{theHalgorithm}{}{}
\makeatother

\usepackage{amsmath}
\usepackage{amssymb}
\usepackage{mathtools}
\usepackage{amsthm}

\usepackage[capitalize,noabbrev]{cleveref}

\usepackage{dsfont}
\usepackage[mathscr]{euscript}
\usepackage{xcolor}
\usepackage{colortbl}
\usepackage{thmtools}
\usepackage{thm-restate}

\usepackage{multirow}
\usepackage{wrapfig}

\usepackage{MnSymbol}
\DeclareMathAlphabet\mathbb{U}{msb}{m}{n}
\usepackage{xpatch}

\def\Nset{\mathbb{N}}
\def\Rset{\mathbb{R}}

\DeclareMathOperator*{\E}{\mathbb E}

\DeclareMathOperator*{\argmin}{argmin}

\DeclareMathOperator{\sign}{sign}

\DeclarePairedDelimiter{\abs}{\lvert}{\rvert} 
\DeclarePairedDelimiter{\bracket}{[}{]}
\DeclarePairedDelimiter{\curl}{\{}{\}}
\DeclarePairedDelimiter{\paren}{(}{)}

\newcommand{\cL}{\mathcal{L}}

\newcommand{\sC}{{\mathscr C}}
\newcommand{\sD}{{\mathscr D}}
\newcommand{\sE}{{\mathscr E}}

\newcommand{\sH}{{\mathscr H}}

\newcommand{\sL}{{\mathscr L}}
\newcommand{\sM}{{\mathscr M}}

\newcommand{\sR}{{\mathscr R}}

\newcommand{\sX}{{\mathscr X}}
\newcommand{\sY}{{\mathscr Y}}

\newcommand{\sfL}{{\mathsf L}}

\newcommand{\1}{\mathsf{1}}

\newcommand{\Rad}{\mathfrak R}

\newcommand{\balpha}{{\boldsymbol \alpha}}
\newcommand{\bbeta}{{\boldsymbol \beta}}
\newcommand{\bgamma}{{\boldsymbol \gamma}}

\newcommand{\hh}{{\sf h}}

\newcommand{\METRO}{\textsc{metro}}

\newcommand{\h}{\widehat}
\newcommand{\ov}{\overline}
\newcommand{\uv}{\underline}

\newcommand{\e}{\epsilon}
\newcommand{\ignore}[1]{}

\newcommand{\ul}{\uv \ell}
\newcommand{\ol}{\ov \ell}

\theoremstyle{plain}
\newtheorem{theorem}{Theorem}[section]

\newtheorem{lemma}[theorem]{Lemma}
\newtheorem{corollary}[theorem]{Corollary}
\theoremstyle{definition}
\newtheorem{definition}[theorem]{Definition}

\theoremstyle{remark}

\newcommand{\algorithmicreturn}{\textbf{return}}
\newcommand{\RETURN}{\algorithmicreturn\ }

\usepackage[disable,textsize=tiny]{todonotes}

\hypersetup{
  breaklinks   = true, 
  colorlinks   = true, 
  urlcolor     = blue, 
  linkcolor    = blue, 
  citecolor   = blue 
}

\usepackage[toc, page, header]{appendix}
\icmltitlerunning{Principled Algorithms for Optimizing
  Generalized Metrics in Binary Classification}

\begin{document}

\twocolumn[
  \icmltitle{Principled Algorithms for Optimizing\\
    Generalized Metrics in Binary Classification}

\begin{icmlauthorlist}
\icmlauthor{Anqi Mao}{courant}
\icmlauthor{Mehryar Mohri}{google,courant}
\icmlauthor{Yutao Zhong}{google}
\end{icmlauthorlist}

\icmlaffiliation{courant}{Courant Institute of Mathematical Sciences,
  New York, NY;}
\icmlaffiliation{google}{Google Research, New York, NY}

\icmlcorrespondingauthor{Anqi Mao}{aqmao@cims.nyu.edu}
\icmlcorrespondingauthor{Mehryar Mohri}{mohri@google.com}
\icmlcorrespondingauthor{Yutao Zhong}{yutaozhong@google.com}

\icmlkeywords{learning theory, consistency, generalized metrics}

\vskip 0.3in
]

\printAffiliationsAndNotice{}

\begin{abstract}

In applications with significant class imbalance or asymmetric costs,
metrics such as the $F_\beta$-measure, AM measure, Jaccard similarity
coefficient, and weighted accuracy offer more suitable evaluation
criteria than standard binary classification loss. However, optimizing
these metrics present significant computational and statistical
challenges.  Existing approaches often rely on the characterization of
the Bayes-optimal classifier, and use threshold-based methods that
first estimate class probabilities and then seek an optimal threshold.
This leads to algorithms that are not tailored to restricted
hypothesis sets and lack finite-sample performance guarantees.
In this work, we introduce principled algorithms for optimizing
generalized metrics, supported by $\sH$-consistency and finite-sample
generalization bounds. Our approach reformulates metric optimization
as a generalized cost-sensitive learning problem, enabling the design
of novel surrogate loss functions with provable $\sH$-consistency
guarantees. Leveraging this framework, we develop new algorithms, \METRO\ (\emph{Metric Optimization}), with
strong theoretical performance guarantees. We report the results
of experiments demonstrating the effectiveness of our methods compared
to prior baselines.

\end{abstract}

\section{Introduction}
\label{sec:intro}

In many applications, performance metrics such as the
\emph{$F_{\beta}$-measure}
\citep{lewis1995evaluating,jansche2005maximum,ye2012optimizing}, the
\emph{AM measure} \citep{menon2013statistical}, the \emph{Jaccard
similarity coefficient} \citep{sokolova2009systematic}, or the
\emph{weighted accuracy} are preferred over the standard zero-one
misclassification loss. These metrics are particularly relevant in
scenarios with significant class imbalance, such as fraud detection,
medical diagnosis, and information retrieval
\citep{lewis1995sequential,drummond2005severe,he2009learning,gu2009evaluation},
or in settings where classification costs are
asymmetrical. However, optimizing these alternative metrics presents
both computational and statistical challenges, as they deviate from
the standard single-loss expectation framework commonly used in
surrogate loss function analysis \citep{steinwart2007compare}.

Research on the design of algorithms for specific instances of such
metrics or the general family has largely focused on the
characterization of the Bayes-optimal classifier. For the standard
zero-one binary classification loss, it is known that the Bayes
classifier can be defined as $x \mapsto \sign(\eta(x) - \frac{1}{2})$,
where $\eta(x)$ is the probability of a positive label conditioned on
$x$.  A similar characterization applies to the weighted zero-one loss
\citep{scott2012calibrated}, where the threshold corresponds to a
weight different from $\frac{1}{2}$.  For imbalanced metrics,
\citet{ye2012optimizing} characterized the Bayes-classifier for the
$F_1$-measure, while \citet{menon2013statistical} characterized the
Bayes classifier for the AM measure as $x \mapsto \sign(\eta(x) -
\delta^*)$ for some $\delta^* \in (0, 1)$.  These results were later
generalized by \citet{KoyejoNatarajanRavikumarDhillon2014} to a
broader family of metrics that can be formulated as a ratio of two
linear functions of true positive (TP), false positive (FP), true
negative (TN), and false negative (FN) statistics. This family of
linear-fractional metrics includes the aforementioned weighted
accuracy, $F_{\beta}$-measures, the AM measure, as well as the
\emph{Jaccard similarity coefficient} \citep{sokolova2009systematic},
among others.

Existing algorithms for optimizing these metrics heavily rely on the
structure of the Bayes-optimal solution
\citep{KoyejoNatarajanRavikumarDhillon2014,
  ParambathUsunierGrandvalet2014}. These methods generally adopt a
two-stage approach: first estimating the conditional probability
$\eta(x)$ and then searching for a suitable threshold $\delta^*$
specific to the metric.

These algorithms naturally come with consistency guarantees
\citep{KoyejoNatarajanRavikumarDhillon2014}. However, consistency is
an asymptotic property and does not provide explicit convergence rate
guarantees. Moreover, it applies only to the overly broad class of all
measurable functions, making it less relevant in practical scenarios
where learning is constrained to a restricted hypothesis class
\citep{long2013consistency,zhang2020bayes}.

Beyond these theoretical limitations, a key drawback of prior work is
its reliance on the structure of the Bayes-optimal classifier, which
may differ significantly from the best predictor within a given
hypothesis class. For example, we show that in a two-dimensional
setting with linear classifiers (see Section~\ref{sec:algo},
Figure~\ref{fig:example}), the best linear classifier derived via a
margin-maximizing classifier can have a significantly different
orientation from the best linear classifier optimized for an
$F_{\beta}$-measure.

\textbf{Contributions.} In contrast to previous work, we introduce a
hypothesis set-specific analysis grounded in recent advances in
\emph{$\sH$-consistency bounds}
\citep*{awasthi2022Hconsistency,awasthi2022multi,mao2023cross,mao2024universal}. Unlike
Bayes-consistency, these bounds are non-asymptotic and explicitly
account for the hypothesis set $\sH$ used in practice. They provide
direct upper bounds on the target estimation error in terms of the
surrogate estimation error, making them more relevant to practical
learning. We leverage this framework to develop new algorithms with
strong theoretical guarantees, including finite-sample learning
bounds.

\textbf{Related work}. A comprehensive discussion on consistency and
generalized metrics in binary classification is provided in
Appendix~\ref{app:related-work}. Here, we briefly summarize the most
relevant prior work.
\citet{KoyejoNatarajanRavikumarDhillon2014} studied a broad family of
performance metrics, including the $F_{\beta}$-measure, AM measure,
Jaccard similarity coefficient, and weighted accuracy. They proposed
two-stage thresholding algorithms that first train a binary classifier
(e.g., logistic regression) to estimate $\eta(x)$ and then optimize a
threshold $\theta$ to maximize the empirical metric of interest. Their
approach comes with a consistency-type guarantee but lacks
finite-sample guarantees.
\citet{ParambathUsunierGrandvalet2014} focused specifically on
optimizing the $F_{1}$-measure. Their algorithm is similar to the
second approach in \citet{KoyejoNatarajanRavikumarDhillon2014} but
additionally considers an extra threshold $\theta'$, both of which are
optimized to maximize the empirical metric. They provide
stability-type guarantees, though their analysis lacks explicit
details.

\textbf{Structure of the paper.} We introduce principled algorithms
for optimizing a broad family of generalized metrics, supported by
$\sH$-consistency bounds and finite-sample generalization bounds.  We
first present an equivalent reformulation of the problem of minimizing
generalized metrics (Section~\ref{sec:formulation}).  This
reformulation allows us to interpret the problem as minimizing a
generalized cost-sensitive target loss function
(Section~\ref{sec:target}).  We then address this broader
cost-sensitive learning problem by introducing a new family of
surrogate loss functions tailored to this framework
(Section~\ref{sec:surrogate}).  We further establish strong
$\sH$-consistency guarantees for these surrogate losses
(Section~\ref{sec:guarantees}). In Section~\ref{sec:algo}, we leverage
these theoretical insights to design new algorithms for optimizing
generalized metrics, \METRO\ (\emph{Metric Optimization}), for which we prove strong performance guarantees.
Finally, we present experimental results in
Section~\ref{sec:experiments}, demonstrating the effectiveness of our
algorithms in comparison to prior baselines.

\section{Preliminaries}
\label{sec:pre}

\textbf{Binary classification}. We consider the familiar setting of
binary classification, where the input space is denoted by $\sX$, the
label space by $\sY = \curl*{+1, -1}$, and the data is distributed
according to an unknown distribution $\sD$ over $\sX \times \sY$. We
will consider prediction functions $h$ mapping from $\sX$ to $\Rset$
and will denote by $\sH_{\rm{all}}$ the family of all measurable
functions of this type. 
For a given loss function $\ell$ mapping from $\sH_{\rm{all}} \times
\sX \times \sY$ to $\Rset$, the \emph{expected loss} of a
hypothesis $h$ and the \emph{best-in-class expected loss} of a
hypothesis set $\sH \subseteq \sH_{\rm all}$ are defined by
$\sE_{\ell}(h) = \E_{(x,y)\sim \sD} \bracket*{\ell(h, x, y)}$ and
$\sE_{\ell}^*(\sH) = \inf_{h \in \sH} \sE_{\ell}(h)$.
The \emph{excess error} of a hypothesis $h$, $\sE_{\ell}(h) -
\sE_{\ell}^*\paren{\sH_{\rm{all}}}$, can be decomposed as the sum of
its \emph{estimation error}, $\sE_{\ell}(h) - \sE_{\ell}^*(\sH)$, and
the approximation error of $\sH$, $\sE_{\ell}^*(\sH) -
\sE_{\ell}^*\paren{\sH_{\rm{all}}}$. Given a sample $S = \paren*{x_1,
  \ldots, x_m}$ and a hypothesis $h$, the \emph{empirical error} is
defined by $\h \sE_{\ell, S}(h) = \frac{1}{m} \sum_{i = 1}^m \ell(h,
x_i, y_i)$.

\textbf{Consistency guarantees}. Given a surrogate loss function
$\ell_1$ and a target loss function $\ell_2$, a fundamental property
of $\ell_1$ with respect to $\ell_2$ is \emph{Bayes-consistency}
\citep{Zhang2003,bartlett2006convexity, steinwart2007compare}.

\begin{definition}[\textbf{Bayes-Consistency}]
A loss function $\ell_1$ is \emph{Bayes-consistent} with respect to a
loss function $\ell_2$ if, for all distributions and sequences
$\{h_n\}_{n\in \Nset}\subset \sH_{\rm all}$, if $[\sE_{\ell_1}(h_n) -
  \sE_{\ell_1}^*\paren*{\sH_{\rm{all}}}]$ tends to zero when $n$ tends
to $+\infty$, then
$[\sE_{\ell_2}(h_n)-\sE_{\ell_2}^*\paren*{\sH_{\rm{all}}} ]$ also
tends to zero.
\end{definition}
While Bayes consistency is a natural and desirable property, it is
inherently asymptotic and applies only to the family of all measurable
functions. As such, it offers no insights into convergence rates or
the behavior of learning algorithms when restricted hypothesis classes
are used, as is typical in machine learning applications.  In this
context, a more informative property is an \emph{$\sH$-consistency
bound} \citep*{awasthi2022Hconsistency,mao2023cross,zhong2025fundamental}, which provides a
quantitative guarantee tailored to specific hypothesis classes.

\begin{definition}[\textbf{$\sH$-Consistency bound}]
A surrogate loss $\ell_1$ admits an \emph{$\sH$-consistency bound}
with respect to the target loss $\ell_2$ if there exists a
non-decreasing concave function $\Gamma \colon \Rset_{+}\to \Rset_{+}$
with $\Gamma(0) = 0$ such that, for all $h\in \sH$ and all
distributions, the following inequality holds:
\begin{multline}
\label{eq:est-bound}
    \sE_{\ell_2}(h) - \sE_{\ell_2}^*(\sH) + \sM_{\ell_2}(\sH)\\
    \leq \Gamma \paren*{\sE_{\ell_1}(h)
      - \sE_{\ell_1}^*(\sH) + \sM_{\ell_1}(\sH)},
\end{multline}
where, for any loss function $\ell$, $\sM_{\ell}(\sH)$ is defined as
$\sM_\ell(\sH) = \sE^*_\ell\paren*{\sH} - \E\bracket*{\inf_{h \in \sH}
  \E\bracket*{\ell(h, x, y) \mid x}}$, and is referred to as the
\emph{minimizability gap}.
\end{definition}
For $\sH = \sH_{\rm{all}}$, the second term coincides with
$\sE^*_\ell\paren*{\sH}$
\citep*{steinwart2007compare,mao2024universal} and the minimizability
gap is zero.  In this case, the bound simplifies to $\sE_{\ell_2}(h) -
\sE_{\ell_2}^*(\sH) \leq \Gamma \paren*{\sE_{\ell_1}(h) -
  \sE_{\ell_1}^*(\sH)}$ which, in particular, implies
Bayes-consistency.
More generally, the minimizability gap vanishes when
$\sE^*_{\ell}(\sH) = \sE_{\ell}^*\paren{\sH_{\rm{all}}}$. The
minimizability gap is always upper bounded by the approximation error
but it can be strictly smaller in many cases.  In general,
$\sH$-consistency bounds provide a stronger, non-asymptotic, and
hypothesis set-dependent consistency guarantee.

\section{Problem Formulation}
\label{sec:formulation}

Our goal is to devise a principled learning algorithm seeking to
optimize any instance within a broad family of generalized metrics. We
first define this family and then reformulate the learning problem
using an alternative loss function.

\subsection{Generalized Metrics} 
\label{sec:generalized-metrics}

We study \emph{generalized metrics} for binary classification, defined
as follows for any $h \in \sH_{\rm{all}}$:
\begin{equation}
\label{eq:target}
\cL(h) = \frac{ \E_{(x, y) \sim \sD}
  \bracket*{ \alpha_1 \hh(x) y
    + \alpha_2 y + \alpha_3 \hh(x) + \alpha_4} }
   { \E_{(x, y) \sim \sD} \bracket*{ \beta_1 \hh(x) y
    + \beta_2 y + \beta_3 \hh(x) + \beta_4}},
\end{equation}
where $\hh(x) = \sign(h(x))$ represents the prediction of a hypothesis
$h$ on an input $x \in \sX$, with $\sign(\alpha) = \1_{\alpha
  \geq 0} - \1_{\alpha < 0}$, and with $\balpha = [\alpha_1,
  \alpha_2, \alpha_3, \alpha_4], \bbeta = [\beta_1, \beta_2, \beta_3,
  \beta_4] \in \Rset^4$.
To avoid notational clutter, we assume the dependence on $(\balpha,
\bbeta)$ is understood from the context and omit it when it is not
necessary.
Note that the true positive (TP), false
positive (FP), true negative (TN), and false negative (FN) statistics can
be expressed in terms of $ \E_{(x, y)} \bracket*{\hh(x) y}$, $ \E_{(x,
  y)} \bracket*{y}$, and $ \E_{(x, y)} \bracket*{\hh(x)}$:
\begin{align*}
  & \text{TP} = \E_{(x, y)} \bracket*{\ov \hh(x) \ov y} && \text{FN}
  = \E_{(x, y)} \bracket*{1- \ov \hh(x) - \ov y + \ov \hh(x) \ov y}\\
 & \text{TN} = \E_{(x, y)} \bracket*{\ov y - \ov \hh(x) \ov y} &&
 \text{FP} = \E_{(x, y)} \bracket*{\ov \hh(x) - \ov \hh(x) \ov y},
\end{align*}
where $\ov \hh(x) = \frac{\hh(x) + 1}{2} \in \curl*{0, 1}$ and $\ov y
= \frac{y + 1}{2} \in \curl*{0, 1}$.
Thus, any ratio of two linear combinations of TP, FP, TN, and FN,
which are considered in \citep{KoyejoNatarajanRavikumarDhillon2014},
can be expressed equivalently in the form of \eqref{eq:target}.  This
formulation covers several widely used metrics, including the AM
measure \citep{menon2013statistical}, the $F_{\beta}$-measure
\citep{ye2012optimizing}, the Jaccard similarity coefficient (JAC)
\citep{sokolova2009systematic}, Weighted Accuracy (WA), and many
others of interest.
Given a hypothesis set $\sH$, our goal is to find a hypothesis $h \in
\sH$ with small loss $\cL(h)$. We denote the best-in-class expected
loss by $\cL^*(\sH) = \inf_{h \in \sH} \cL(h)$. Given a sample $S =
\paren*{x_1, \ldots, x_m}$ and a hypothesis $h$, the \emph{empirical
loss} is defined by $\h \cL_{S}(h) = \frac{ \frac{1}{m} \sum_{i = 1}^m
  \bracket*{ \alpha_1 \hh(x_i) y_i + \alpha_2 y_i + \alpha_3 \hh(x_i)
    + \alpha_4} }{ \frac1m \sum_{i = 1}^m \bracket*{ \beta_1 \hh(x_i)
    y_i + \beta_2 y_i + \beta_3 \hh(x_i) + \beta_4}}$. 

\subsection{Equivalent Problem}
\label{sec:equivalent-problem}

The generalized metrics introduced above are defined as the ratio of
two expected loss functions. This formulation differs from the more
familiar single-loss expectations, which are commonly used for
deriving and analyzing surrogate loss functions in machine
learning. In this section, we present an equivalent reformulation of
the problem of minimizing $\cL(h)$. This
reformulation will provide a more convenient framework for designing
surrogate loss functions.

For any $(h, x, y) \in \sH_{\rm{all}} \times \sX \times \sY$, we
define two loss functions: $\ell_{\balpha} \colon (h, x, y) \mapsto
\alpha_1 \hh(x) y + \alpha_2 y + \alpha_3 \hh(x) + \alpha_4$ and
$\ell_{\bbeta} \colon (h, x, y) \mapsto \beta_1 \hh(x) y + \beta_2 y +
\beta_3 \hh(x) + \beta_4$.  Using these definitions, $\cL(h)$ can be
rewritten as follows:
\begin{equation*}
  \cL(h)
  = \frac{\E_{(x, y) \sim \sD}
    \bracket*{ \ell_{\balpha}(h, x, y) }}{\E_{(x, y) \sim \sD}
    \bracket*{ \ell_{\bbeta}(h, x, y) }},
\end{equation*}
that is, the fractional form of the expected losses of
$\ell_{\balpha}$ and $\ell_{\bbeta}$. Next, to minimize
$\cL(h)$, we reformulate it as an equivalent
optimization problem, which will facilitate further analysis and
surrogate design.

For any $\lambda$ (and $\balpha$ and $\bbeta$), define the loss
function $\ell^{\lambda}$ by:
\begin{equation}
\label{eq:target-equiv}
\forall (h, x, y), \, \ell^{\lambda}(h, x, y)
=  \ell_{\balpha}(h, x, y) - \lambda \ell_{\bbeta}(h, x, y).
\end{equation}
We will denote by $\h \sE_{\ell^\lambda, S}$ the empirical loss of
$\ell^\lambda$ over a sample $S$. Without loss of generality, we
assume throughout that $\E_{(x, y) \sim \sD}
\bracket*{\ell_{\bbeta}(h, x, y)}$ is positive for all $h \in \sH$. If
this condition does not hold, it can be enforced by redefining
$\ell_{\bbeta} = -\ell_{\bbeta}$ and $\ell_{\balpha} =
-\ell_{\balpha}$, as needed.  For convenience, we will further define
$\ul_\beta = \inf_{h \in \sH} \E_{(x, y) \sim \sD}[\ell_\beta(h, x,
  y)]$ and assume $\ul_\beta > 0$. Similarly, let $\ol_\beta = \sup_{h
  \in \sH} \E_{(x, y) \sim \sD}[\ell_\beta(h, x, y)]$, where
$\ol_\beta < +\infty$.  The following theorem establishes that
minimizing $\cL(h)$ over $\sH$ is equivalent to minimizing
$\sE_{\ell^{\lambda^*\!\!}}(h)$ over $\sH$, where $\lambda^* =
\cL^*(\sH)$.

\begin{restatable}{theorem}{Equiv}
\label{thm:equiv}
The equality $\cL(h^*) = \cL^*(\sH)$ holds for $h^* \in \sH$ if and
only if $\sE_{\ell^{\lambda^*\!\!}}(h^*) = \sE_{\ell^{\lambda^*\!\!}}(\sH) = 0$.
\end{restatable}
\begin{proof}
Assume that there exists a hypothesis $h^* \in \sH$ such that
$\cL(h^*) = \cL^*(\sH)$ holds. We
have for all $h \in \sH$:
\begin{align*}
  \lambda^*
  = \cL(h^*) & =  \frac{\E
    \bracket*{ \ell_{\balpha}(h^*, x, y) }}{\E
    \bracket*{ \ell_{\bbeta}(h^*, x, y) }}
  \leq \frac{\E
    \bracket*{ \ell_{\balpha}(h, x, y) }}
  {\E \bracket*{ \ell_{\bbeta}(h, x, y) }}.
\end{align*}
Thus, under the assumption, the following holds for all $h \in \sH$:
\begin{align*}
  \lambda^* \E_{(x, y) \sim \sD} \bracket*{ \ell_{\bbeta}(h, x, y) }
  & \leq \E_{(x, y) \sim \sD} \bracket*{ \ell_{\balpha}(h, x, y) },\\
  \lambda^* \E_{(x, y) \sim \sD} \bracket*{ \ell_{\bbeta}(h^*, x, y) }
  & = \E_{(x, y) \sim \sD} \bracket*{ \ell_{\balpha}(h^*, x, y) }.
\end{align*}
This implies that $\sE_{\ell^{\lambda^*\!\!}}(h^*)  = \sE_{\ell^{\lambda^*\!\!}}(\sH) = 0$, which completes one direction of the proof.

Assume now that there exists $h^* \in \sH$ such that
$\sE_{\ell^{\lambda^*\!\!}}(h^*) =
\sE_{\ell^{\lambda^*\!\!}}(\sH) = 0$ holds. We have for
all $h \in \sH$:
\begin{align*}
& \E_{(x, y) \sim \sD} \bracket*{\ell_{\balpha}(h^*, x, y)} - \lambda^* \E_{(x, y) \sim \sD} \bracket*{ \ell_{\bbeta}(h^*, x, y)} = 0\\
& \E_{(x, y) \sim \sD} \bracket*{\ell_{\balpha}(h, x, y)} - \lambda^* \E_{(x, y) \sim \sD} \bracket*{ \ell_{\bbeta}(h, x, y)} \geq 0.
\end{align*}
Thus, we have $\lambda^* = \frac{\E_{(x, y) \sim
    \sD} \bracket*{ \ell_{\balpha}(h^*, x, y) }}{\E_{(x, y) \sim \sD}
  \bracket*{ \ell_{\bbeta}(h^*, x, y) }} = \cL(h^*)
\leq \cL(h)$ for all $h \in \sH$, which implies 
$\cL(h^*) = \cL^*(\sH)$. This
completes the proof.
\end{proof}

More generally, the following non-asymptotic equivalence holds. The
proof is given in Appendix~\ref{app:equiv-non}.
\begin{restatable}{theorem}{EquivNon}
  \label{thm:equiv-non}
Fix $\eta \geq 0$ and $h \in \sH$. Then, the inequality
$\sE_{\ell^{\lambda^*}}(h) \leq \eta$ holds if and
only if $\cL(h) - \cL^*(\sH) \leq
\frac{\eta}{\E_{(x, y) \sim \sD} \bracket*{ \ell_{\bbeta}(h, x, y)
}}$.
\end{restatable}

Theorems~\ref{thm:equiv} and \ref{thm:equiv-non} show that our problem
can be reduced to minimizing the expected value of the loss function
$\ell^{\lambda^*\!\!}$. However, this remains
intractable because $\ell^{\lambda^*\!\!}$ is non-differentiable and even
non-continuous as a function of $h$ (as it is linear in $\hh$). In the
next section, we will define a general family of consistent surrogate
losses for $\ell^{\lambda^*\!\!}$ that are more suitable for practical
optimization techniques.

\section{General Cost-Sensitive Learning}
\label{sec:csl}

In this section, we first establish that $\ell^{\lambda^*\!\!}$ can be
interpreted as a generalized cost-sensitive target loss function.  We
then consider the broader problem of cost-sensitive learning and
introduce a novel family of surrogate loss functions tailored to this
framework. Finally, we present strong theoretical guarantees for these
surrogate losses, demonstrating that minimizing them yields efficient
algorithms for both general cost-sensitive learning and the
optimization of $\ell^{\lambda^*\!\!}$.

\subsection{General Target Loss}
\label{sec:target}

Define $\bgamma = \bracket*{\gamma_1, \gamma_2, \gamma_3, \gamma_4}$,
where $\gamma_i = \alpha_i - \lambda^* \beta_i$, for $i \in \curl*{1,
  2, 3, 4}$. Then, for any $h \in \sH$ and $(x, y) \in \sX \times \sY$,
$\ell^{\lambda^*\!\!}(h, x, y)$ can be expressed as:
\begin{align}
\label{eq:target-equiv-2}
\gamma_1 \hh(x) y  + \gamma_2 y + \gamma_3 \hh(x) + \gamma_4
= \sfL_{\bgamma}(\hh(x), y), 
\end{align}
where $\sfL_{\bgamma}$ is a cost-sensitive function over $\sY
\times \sY$ defined as
\[\sfL_{\bgamma}(y', y) = \begin{cases}
\gamma_1 + \gamma_2 + \gamma_3 + \gamma_4 & y' = 1, y = 1,\\
-\gamma_1 - \gamma_2 + \gamma_3 + \gamma_4 & y' = 1, y = -1,\\
-\gamma_1 + \gamma_2 - \gamma_3 + \gamma_4 & y' = -1, y = 1,\\
\gamma_1 - \gamma_2 - \gamma_3 + \gamma_4 & y' = -1, y = -1.
\end{cases}\]

Adding a constant to $\sfL_{\bgamma}$ does not affect its
minimization.  Therefore, we can augment it with a non-negative
constant $\tau$ to ensure $\sfL_{\bgamma} + \tau \geq 0$. A suitable
choice is $\tau = \abs*{\gamma_1} + \abs*{\gamma_2} + \abs*{\gamma_3}
+ \abs*{\gamma_4}$.

This can be regarded as a special case of the \emph{general
cost-sensitive learning} problem \citep{elkan2001foundations}. In this
context, we consider a cost-sensitive loss function $\ov \sfL \colon
\curl*{+1, -1} \times \curl*{+1, -1} \to \Rset_{+}$, which defines
four non-negative costs: $\ov \sfL(+1, +1)$, $\ov \sfL(+1, -1)$, $\ov
\sfL(-1, +1)$, and $\ov \sfL(-1, -1)$. These costs depend on the
prediction $\hh(x) \in \curl*{-1, +1}$ and the true label $y \in
\curl*{-1, +1}$. We denote by $\sfL$ the target loss function induced
by $\ov \sfL$, defined as:
\begin{equation} 
\label{eq:csl}
\forall (h, x, y), \quad \sfL(h, x, y) = \ov \sfL(\hh(x), y).
\end{equation}
This approach is referred to as general cost-sensitive learning, as
opposed to the specific cost-sensitive learning problem analyzed in
\citep{scott2012calibrated}, where $\ov\sfL(+1, -1) = \theta$,
$\ov\sfL(-1, +1) = 1 - \theta$, and $\ov \sfL(+1, +1) = \ov \sfL(-1,
-1) = 0$. Furthermore, the existing $\theta$-weighted surrogate losses
in \citep{scott2012calibrated} are not applicable in this general
setting.

\subsection{General Surrogate Losses}
\label{sec:surrogate}

\begin{table*}[t]
\caption{Common Margin-Based Losses and Their General Cost-Sensitive Surrogate Extensions.}
  \label{tab:sur}
  \centering
  \vskip .1in
  \resizebox{.75\textwidth}{!}{
  \begin{tabular}{@{\hspace{0cm}}lll@{\hspace{0cm}}}
    \toprule
      Name & $\Phi(t)$ & Cost-Sensitive Surrogate Loss $\sfL_{\Phi}$\\
    \midrule
     Exponential & $\Phi_{\rm{exp}}(t) = e^{-t}$ & $\ov \sfL(+1, y) e^{h(x)} + \ov \sfL(-1, y) e^{-h(x)}$    \\
     Logistic & $\Phi_{\rm{log}}(t) = \log\paren*{1 + e^{-t}}$ & $\ov \sfL(+1, y) \log\paren*{1 + e^{h(x)}} + \ov \sfL(-1, y) \log\paren*{1 + e^{-h(x)}}$ \\
     Quadratic & $\Phi_{\rm{quad}}(t) = \max\curl*{1 - t, 0}^2$ & $\ov \sfL(+1, y) \Phi_{\rm{quad}}(-h(x)) + \ov \sfL(-1, y) \Phi_{\rm{quad}}(h(x))$ \\
     Hinge & $\Phi_{\rm{hinge}}(t) = \max\curl*{1 - t, 0}$ & $\ov \sfL(+1, y) \Phi_{\rm{hinge}}(-h(x)) + \ov \sfL(-1, y) \Phi_{\rm{hinge}}(h(x))$ \\
     Sigmoid & $\Phi_{\rm{sig}}(t) = 1 - \tanh(k t), k > 0$ & $\ov \sfL(+1, y) \Phi_{\rm{sig}}(-h(x)) + \ov \sfL(-1, y) \Phi_{\rm{sig}}(h(x))$  \\
     $\rho$-Margin & $\Phi_{\rho}(t) = \min\curl*{1,  \max\curl*{0, 1-\frac{t}{\rho}}}, \rho > 0$ & $\ov \sfL(+1, y) \Phi_{\rho}(-h(x)) + \ov \sfL(-1, y) \Phi_{\rho}(h(x))$ \\ 
    \bottomrule
  \end{tabular}
  }
\end{table*}

As with many target loss functions in learning problems, such as the
zero-one loss in binary classification, directly minimizing the
general cost-sensitive loss function $\sfL$ is intractable for most
hypothesis sets due to its non-continuity and
non-differentiability. Instead, surrogate losses are typically adopted
in practice. These surrogate losses are designed to be consistent, or
even $\sH$-consistent in the standard classification settings
\citep{Zhang2003,bartlett2006convexity,awasthi2022Hconsistency,
  awasthi2022multi,mao2023cross}. They are frequently
formulated as margin-based loss functions \citep{lin2004note}, which
are defined by non-increasing functions $\Phi \colon t \to \Rset$ that
upper bound $t \mapsto 1_{t \leq 0}$. For example, $\Phi$ could be the
hinge loss, $t \mapsto \max\curl*{0, 1 - t}$, or the logistic loss, $t
\mapsto \log(1 + e^{-t})$.

Here, we define the following surrogate loss functions for the general
cost-sensitive loss function $\sfL$ by extending the standard
margin-based loss function to the general cost-sensitive setting:
\begin{equation}
\label{eq:sur}
\sfL_{\Phi}(h, x, y) = \ov \sfL(+1, y) \Phi(-h(x))
+ \ov \sfL(-1, y) \Phi(h(x)).
\end{equation}
Table~\ref{tab:sur} lists common examples of $\Phi$ along with the
corresponding general cost-sensitive surrogate losses. A special case
of \eqref{eq:sur} arises when the costs satisfy $\ov\sfL(+1, -1) =
\theta$, $\ov\sfL(-1, +1) = 1 - \theta$, and $\ov \sfL(+1, +1) = \ov
\sfL(-1, -1) = 0$. This corresponds to the $\theta$-weighted surrogate
loss considered in
\citep{scott2012calibrated,KoyejoNatarajanRavikumarDhillon2014}. Our
formulation generalizes and significantly extends this surrogate loss
framework to address the broader context of general cost-sensitive
learning.

It is important to highlight that our proposed algorithm, detailed in
Section~\ref{sec:algo}, for optimizing generalized metrics differs
fundamentally from the second algorithm introduced in
\citep{KoyejoNatarajanRavikumarDhillon2014} (see also
\citep{ParambathUsunierGrandvalet2014}), despite both leveraging a
sub-algorithm for cost-sensitive learning.
As discussed in Section~\ref{sec:intro}, their approach involves
approximating the Bayes-classifier, a threshold function, using the
$\theta$-weighted cost-sensitive surrogate loss function.  In
contrast, our algorithm minimizes a \emph{general} cost-sensitive
surrogate loss function that is $\sH$-consistent (see
Section~\ref{sec:guarantees}) with respect to a \emph{general}
cost-sensitive target loss function \eqref{eq:target-equiv-2}, with
label-dependent costs that take into account the best-in-class error
of the generalized metric in binary classification, which can be
approximated through a binary search-based algorithm (see
Section~\ref{sec:algo}).

\subsection{Theoretical Guarantees}
\label{sec:guarantees}

In this section, we establish strong theoretical guarantees for a
surrogate loss $\sfL_{\Phi}$. Specifically, we derive
$\sH$-consistency bounds for $\sfL_{\Phi}$ with respect to the
cost-sensitive loss function $\sfL$, focusing on commonly used
hypothesis sets.

We define a hypothesis set $\sH$ as \emph{regular} if, for any $x \in
\sX$, the set of predictions made by the hypotheses in $\sH$ on $x$
covers all possible labels: $\curl*{\hh(x) \colon h \in \sH}
= \curl*{+1, -1}$. Commonly used hypothesis sets, such as linear
models, neural networks, and the family of all measurable functions,
all naturally satisfy this regularity condition.

It was shown by \citet*{awasthi2022Hconsistency} that common
margin-based loss functions, such as the hinge loss, logistic loss,
and exponential loss, admit strong $\sH$-consistency bounds with
respect to the binary zero-one loss function $\ell_{0-1} \colon (h, x,
y) \mapsto \1_{\hh(x) \neq y}$ when using such regular
hypothesis sets. The next result shows that, for such margin-based
loss functions $\Phi$, their corresponding cost-sensitive surrogate
losses $\sfL_{\Phi}$ (Eq.~\eqref{eq:sur}) also admit
$\sH$-consistency bounds with respect to the cost-sensitive loss
$\sfL$ (Eq.~\eqref{eq:csl}).

\begin{restatable}{theorem}{Bound}
\label{thm:bound}
Assume that $\sfL$ takes values in $[0, \sfL_{\max}]$.  Let $\sH$ be a
regular hypothesis set and $\Phi$ a margin-based loss function for the
binary zero-one loss function $\ell_{0-1}$. Assume that $\Phi$
admits a $\Gamma$-$\sH$-consistency bound with respect to $\ell_{0-1}$
for a function $\Gamma\colon t \mapsto \beta\, t^{\alpha}$, with
$\alpha \in (0, 1]$ and $\beta > 0$. Then, $\sfL_{\Phi}$
  admits a $\ov \Gamma$-$\sH$-consistency bound with respect to
$\sfL$, where $\ov \Gamma(t)
= \beta  \paren*{2 \, \sfL_{\max}}^{1 - \alpha} t^{\alpha}$.
\ignore{
  the following $\sH$-consistency bound
  holds for all $h\in \sH$ and all distributions:
  \ifdim\columnwidth=\textwidth {
\begin{equation*}
  \sE_{\ell_{0-1}}(h) - \sE^*_{\ell_{0-1}}(\sH) + \sM_{\ell_{0-1}}(\sH)
  \leq \Gamma \paren*{\sE_{\Phi}(h)
    - \sE^*_{\Phi}(\sH) + \sM_{\Phi}(\sH)}.
\end{equation*}
}\else{
\begin{multline*}
  \sE_{\ell_{0-1}}(h) - \sE^*_{\ell_{0-1}}(\sH) + \sM_{\ell_{0-1}}(\sH)\\
  \leq \Gamma \paren*{\sE_{\Phi}(h)
    - \sE^*_{\Phi}(\sH) + \sM_{\Phi}(\sH)}.
\end{multline*}
}\fi
Then, the following $\sH$-consistency bound holds for all $h\in \sH$
and all distributions:
\ifdim\columnwidth=\textwidth {
\begin{equation*}
\sE_{\sfL}(h) - \sE_{\sfL}^*(\sH) + \sM_{\sfL}(\sH)
\leq \ov \Gamma \paren*{\sE_{\sfL_{\Phi}}(h)
  - \sE_{\sfL_{\Phi}}^*(\sH) + \sM_{\sfL_{\Phi}}(\sH)},    
\end{equation*}
}\else
{
\begin{multline*}
\sE_{\sfL}(h) - \sE_{\sfL}^*(\sH) + \sM_{\sfL}(\sH)\\
\leq \ov \Gamma \paren*{\sE_{\sfL_{\Phi}}(h)
  - \sE_{\sfL_{\Phi}}^*(\sH) + \sM_{\sfL_{\Phi}}(\sH)},
\end{multline*}
}\fi
where $\ov \Gamma(t)
= \beta  \paren*{2 \, \sfL_{\max}}^{1 - \alpha} t^{\alpha}$.
}
\end{restatable}
The proof is included in Appendix~\ref{app:bound}.
Based on the results of \citet{awasthi2022Hconsistency}, the theorem
holds with $\Gamma(t) = \sqrt{2t}$ for the logistic loss and
the exponential loss ($\alpha = 1/2$), $\Gamma(t) = \sqrt{t}$ for the
quadratic loss, and $\Gamma(t) = t$ for the hinge loss, sigmoid loss and $\rho$-margin loss ($\alpha = 1$).

As already mentioned, when the best-in-class error coincides with the
Bayes error, $\sE^*_{\ell}(\sH) = \sE_{\ell}^*\paren{\sH_{\rm{all}}}$
for $\ell = \sfL_{\Phi}$ and $\ell = \sfL$, the minimizability gaps
$\sM_{\sfL}(\sH)$ and $\sM_{\sfL_{\Phi}}(\sH)$ vanish. Under these
conditions, the $\sH$-consistency bound guarantees that when the
surrogate estimation error $\sE_{\sfL_{\Phi}}(h) -
\sE_{\sfL_{\Phi}}^*(\sH)$ is reduced to $\e$, the estimation error of
the cost-sensitive loss $ \sE_{\sfL}(h) - \sE_{\sfL}^*(\sH)$ is upper
bounded by $\ov \Gamma(\e)$.

More generally, since a concave function $\Gamma$ is sub-additive
over $\Rset_+$, the following guarantee holds:
\begin{align*}
& \sE_{\sfL}(h) - \sE_{\sfL}^*(\sH))\\
  & \leq \ov \Gamma \paren*{\sE_{\sfL_{\Phi}}(h)
    - \sE_{\sfL_{\Phi}}^*(\sH)}
+ \ov \Gamma\paren*{\sM_{\sfL_{\Phi}}(\sH)} - \sM_{\sfL}(\sH).
\end{align*}
When the minimizability gaps (or the upper bounding approximation
errors) are small, the last terms, $\bracket*{\ov
  \Gamma\paren*{\sM_{\sfL_{\Phi}}(\sH)} - \sM_{\sfL}(\sH)}$ is
also small and close to zero.
In particular, when $\sH = \sH_{\rm{all}}$, the family of all
measurable functions, all minimizability gap terms in
Theorem~\ref{thm:bound} vanish, yielding the following result.

\begin{corollary}
\label{cor:bound}
Fix a margin-based loss function $\Phi$. Assume that there exists a
function $\Gamma(t) = \beta\, t^{\alpha}$ for some $\alpha \in (0, 1]$
  and $\beta > 0$, such that the following excess error bound holds
  for all $h \in \sH_{\rm{all}}$ and all distributions:
\begin{equation*}
  \sE_{\ell_{0-1}}(h) - \sE^*_{\ell_{0-1}}\paren{\sH_{\rm{all}}}
  \leq \Gamma \paren*{\sE_{\Phi}(h)
    - \sE^*_{\Phi}\paren{\sH_{\rm{all}}}}.
\end{equation*}
Then, the following excess error bound holds for all $h\in
\sH_{\rm{all}}$ and all distributions:
\begin{equation*}
  \sE_{\sfL}(h) - \sE_{\sfL}^*\paren{\sH_{\rm{all}}}
  \leq \ov \Gamma \paren*{\sE_{\sfL_{\Phi}}(h)
    - \sE_{\sfL_{\Phi}}^*\paren{\sH_{\rm{all}}}},    
\end{equation*}
where $\ov \Gamma(t)
= \beta  \paren*{2 \, \sfL_{\max}}^{1 - \alpha} t^{\alpha}$.
\end{corollary}
Building on the results of \citet{awasthi2022Hconsistency} for
$\Gamma(t)$ already mentioned and Corollary~\ref{cor:bound}, we now
derive the following result.
\begin{corollary}
\label{cor:bound-example}
For all $h\in \sH_{\rm{all}}$ and all distributions,
\begin{equation*}
    \sE_{\sfL}(h) - \sE_{\sfL}^*\paren{\sH_{\rm{all}}}
    \leq 
    \ov \Gamma\paren*{\sE_{\sfL_{\Phi}}(h)
      - \sE_{\sfL_{\Phi}}^*\paren{\sH_{\rm{all}}}},
\end{equation*}
where $\ov \Gamma (t) = 2 \sqrt{\sfL_{\max} \, t}$
for $\Phi = \Phi_{\rm{exp}}$ and $\Phi_{\rm{log}}$, $\ov \Gamma (t) =
\sqrt{2 \, \sfL_{\max} \, t}$ for $\Phi =
\Phi_{\rm{quad}}$, and $\ov \Gamma (t) = t$ for $\Phi =
\Phi_{\rm{hinge}}$, $\Phi_{\rm{sig}}$, and $\Phi_{\rho}$.
\end{corollary}
By taking the limit on both sides, we establish the Bayes-consistency
of these cost-sensitive surrogate losses $\sfL_{\Phi}$ with
respect to the cost-sensitive target loss $\sfL$. More generally,
Corollary~\ref{cor:bound} demonstrates that $\sfL_{\Phi}$ admits
an excess error bound with respect to $\sfL$ if $\Phi$ admits an
excess error bound with respect to $\ell_{0-1}$.

\section{Algorithm for Generalized Metrics}
\label{sec:algo}

In this section, we build on the previous theoretical analysis to
develop algorithms for optimizing general metrics with strong
guarantees. We first characterize $\lambda^*$, motivating a binary
search algorithm under oracle access to the sign of the expected
loss. We then propose an algorithm based on empirical minimization of
a surrogate loss $\sfL_\Phi$ for $\ell^\lambda$ and introduce a
simpler cross-validation approach for selecting $\lambda$. Finally, we
discuss the theoretical foundations of our algorithms, compare them to
prior work, and highlight cases where existing methods may fail due to
reliance on the Bayes-optimal solution.

First, note that in the general cost-sensitive learning problem, the costs $\ov
\sfL(+1, y)$ and $\ov \sfL(-1, y)$ are known \emph{a priori}. In our
scenario, the costs defining $\ell^{\lambda^*\!\!}$
depend on $\lambda^*$, which is not known.
We will seek to determine or approximate $\lambda^*$. 
Recall that $\lambda^* = \cL^*(\sH) = \inf_{h \in
  \sH} \frac{\E_{(x, y) \sim \sD} \bracket*{ \ell_{\balpha}(h, x, y)
}}{\E_{(x, y) \sim \sD} \bracket*{ \ell_{\bbeta}(h, x, y) }}$ and that
the expected loss of $h \in \sH$ with respect to the loss
function $\ell^{\lambda}$ can be expressed as
follows:
\begin{equation*}
  \sE_{\ell^\lambda}(h) = \E_{(x, y) \sim \sD}
  \bracket*{\ell_{\balpha}(h, x, y)} - \lambda \E_{(x, y) \sim \sD}
  \bracket*{ \ell_{\bbeta}(h, x, y)}.
\end{equation*}
The following provides a key characterization of $\lambda^*$.

\begin{restatable}{theorem}{LambdaStar}
\label{thm:lambda-star}
We have $\sE_{\ell^{\lambda^*\!\!}}^*(\sH) = 0$ and, for
any $\lambda \in \Rset$, 
$\sign\paren[\big]{\sE_{\ell^\lambda}^*(\sH)} =
\sign(\lambda^* - \lambda)$.
\end{restatable}
The proof is included in
Appendix~\ref{app:lambda-star}. Theorem~\ref{thm:lambda-star}
provides a characterization of the sign of the best-in-class
expected loss $\sE_{\ell^\lambda}^*(\sH)$
in terms of the sign of $\lambda^* - \lambda$.

This naturally suggests a binary search-based algorithm to compute an
$\e$-approximation of $\lambda^*$, assuming oracle access to the sign
of $\sE_{\ell^\lambda}^*(\sH)$.  The pseudocode of this algorithm is
provided in Algorithm~\ref{alg:binary-search},where $\lambda_{\min}$
and $\lambda_{\max}$ denote the minimum and maximum possible values of
$\lambda$, respectively.  These bounds can be determined from the
range of $\lambda^*$ using the formulation $\eqref{eq:target}$ for the
given pair $\paren*{\balpha, \bbeta}$.

\begin{algorithm}[t]
\caption{Binary search estimation of $\lambda^*$}
\label{alg:binary-search}
\begin{algorithmic}[1]
  \INPUT $\e$
  \STATE Initialize $[a, b] \gets \bracket*{\lambda_{\min},
    \lambda_{\max}}$
   \REPEAT
   \STATE  $\lambda \gets \frac{a + b}{2}$
      \IF{($\sE_{\ell^\lambda}^*(\sH) > 0$)}
         \STATE $[a, b] = [\lambda, b]$
      \ELSE
         \STATE $[a, b] = [a, \lambda]$
      \ENDIF
      \UNTIL{$\abs*{b - a} \leq \e$}
  \STATE \RETURN $\lambda$
\end{algorithmic}
\end{algorithm}

\begin{restatable}{theorem}{Rate}
\label{thm:rate}
Let $\e > 0$ be fixed. Algorithm~\ref{alg:binary-search} returns an
$\e$-approximation $\lambda$ of $\lambda^*$ in $O
\paren*{\log_2\paren*{\frac{\lambda_{\max} - \lambda_{\min}}{\e}}}$ time,
such that the following property holds:
\[
\sE_{\ell^\lambda}^*(\sH) \leq \sE_{\ell^{\lambda^*\!\!}}^*(\sH) + \e \ol_\beta = \e \ol_\beta.
\]\\[-1.1cm]
\end{restatable}
The proof is presented in Appendix~\ref{app:rate}.
Of course, in practice, we do not have oracle access to the sign of
$\sE_{\ell^\lambda}^*(\sH)$.  However, we can approximate
$\sE_{\ell^\lambda}^*(\sH)$ by computing the solution $\h h_S$ that
minimizes a surrogate loss $\sfL_{\Phi}$ of $\ell^\lambda$ on a
labeled sample $S$ of size $m$. To do that, we first provide a
generalization bound for the target loss $\ell^{\lambda}$ by using our
$\sH$-consistency bounds presented in
Section~\ref{sec:guarantees}. Given a sample $S$ of size $m$, we
denote by $\Rad_m^\lambda(\sH)$ the Rademacher complexity of the
function class $\curl*{(x, y) \mapsto \sfL_{\Phi}(h, x, y) \colon h
  \in \sH}$ and $B_{\lambda} = \sup_{h, x, y }\sfL_{\Phi}(h, x, y)$
an upper bound on $\sfL_{\Phi}$.

\begin{restatable}{theorem}{Generalization}
\label{thm:generalization}
Assume that the surrogate loss $\sfL_{\Phi}$ admits a $\ov
\Gamma$-$\sH$-consistency bound with respect to
$\ell^{\lambda}$.
Then, for any $\delta > 0$, with probability at least $1 - \delta$
over the draw of a sample $S$ from $\sD^m$, the following estimation
bound holds for an empirical minimizer $\h h_S \in \sH$ of the
$\sfL_{\Phi}$ over $S$:
\ifdim\columnwidth= \textwidth
{
\begin{equation*}
\sE_{\ell^\lambda}(\h h_S) - \sE_{\ell^\lambda}^*(\sH)
\leq \Gamma
\paren[\bigg]{\sM_{\sfL_{\Phi}}(\sH) + 4 \Rad_m^\lambda(\sH)
  + 2 B_\lambda \sqrt{\tfrac{\log \frac{2}{\delta}}{2m}}}
- \sM_{\ell^\lambda}(\sH).
\end{equation*}
}\else
{
\begin{multline*}
\sE_{\ell^\lambda}(\h h_S) - \sE_{\ell^\lambda}^*(\sH)\\
\leq \ov \Gamma
  \paren[\bigg]{\sM_{\sfL_{\Phi}}(\sH) + 4 \Rad_m^\lambda(\sH) +
2 B_\lambda \sqrt{\tfrac{\log \frac{2}{\delta}}{2m}}}
\!-\! \sM_{\ell^\lambda}(\sH).
\end{multline*}
}\fi \\[-1cm]
\end{restatable}
Note that such $\ell^\lambda$-estimation loss guarantees for the
minimizer of a surrogate loss $\sfL_\Phi$ can rarely be found in the
literature.  The proof is presented in
Appendix~\ref{app:generalization}.  For
Theorem~\ref{thm:generalization}, the parameter $\lambda$ is fixed.
But, the bound of the theorem can be generalized to hold uniformly for
all $\lambda \in \bracket*{\lambda_{\min}, \lambda_{\max}}$ by
covering the interval using sub-intervals of size $1/m$. 

\begin{restatable}{theorem}{GeneralizationUniform}
\label{thm:generalization-uniform}
Assume that the surrogate loss $\sfL_{\Phi}$ admits a $\ov
\Gamma$-$\sH$-consistency bound with respect to
$\ell^{\lambda}$.
Then, for any $\delta > 0$,  with probability at least $1 - \delta$ over the draw of a sample
$S$ from $\sD^m$, the following estimation bound holds for an
empirical minimizer $\h h_S \in \sH$ of the surrogate loss
$\sfL_{\Phi}$ over $S$ and $\lambda \in [\lambda_{\min},
  \lambda_{\max}]$:
  \ifdim\columnwidth= \textwidth
{
\begin{equation*}
\sE_{\ell^\lambda}(\h h_S) - \sE_{\ell^\lambda}^*(\sH)
\leq \ov \Gamma
\paren[\bigg]{\sM_{\sfL_{\Phi}}(\sH) + 4\Rad_m^{\lambda}(\sH) + \frac{8 \max_{i} \abs*{\beta_i} B_{\Phi} }{m^2}
  + \bracket*{2 B_{\lambda} + \frac{4 \max_{i} \abs*{\beta_i} B_{\Phi} }{m}}\sqrt{\tfrac{\log \frac{2 \Delta \lambda \, m}{\delta}}{2m}}} + \frac{\ol_\beta}{m}.
\end{equation*}
}\else
{
\begin{multline*}
\sE_{\ell^\lambda}(\h h_S) - \sE_{\ell^\lambda}^*(\sH)
\leq \ov \Gamma
\paren[\bigg]{\sM_{\sfL_{\Phi}}(\sH) + 4\Rad_m^{\lambda}(\sH)\\
+ \frac{8 \max_{i} \abs*{\beta_i} B_{\Phi} }{m^2}
+ \bracket*{2 B_{\lambda} + \frac{4 \max_{i} \abs*{\beta_i} B_{\Phi} }{m}}
\sqrt{\tfrac{\log \frac{2 \Delta \lambda \, m}{\delta}}{2m}}}
+ \frac{\ol_\beta}{m},
\end{multline*}
}\fi
where $\Delta \lambda = \lambda_{\max} - \lambda_{\min}$.
In particular, when $\ov \Gamma (t) = 2 \paren*{ \sfL_{\max}}^{\frac{1}{2}} t^{\frac{1}{2}}$
for $\Phi = \Phi_{\rm{log}}$, the bound can be expressed as follows:
\ifdim\columnwidth= \textwidth
{
\begin{equation*}
\sE_{\ell^\lambda}(\h h_S) - \sE_{\ell^\lambda}^*(\sH)
\leq \paren*{2 \, \sfL_{\max}}^{\frac{1}{2}}
\bracket[\bigg]{\sM_{\sfL_{\Phi}}(\sH) + 4\Rad_m^{\lambda}(\sH) + \frac{8 \max_{i} \abs*{\beta_i} B_{\Phi} }{m^2}
  + \bracket[\bigg]{2 B_{\lambda} + \frac{4 \max_{i} \abs*{\beta_i} B_{\Phi} }{m}}\sqrt{\tfrac{\log \frac{2 (\Delta \lambda) \, m}{\delta}}{2m}}}^{\frac{1}{2}}
\mspace {-10mu} + \frac{\ol_\beta}{m}.
\end{equation*}
}\else
{
\begin{multline*}
\sE_{\ell^\lambda}(\h h_S) - \sE_{\ell^\lambda}^*(\sH)
\leq \paren*{2 \, \sfL_{\max}}^{\frac{1}{2}}
\bracket[\bigg]{\sM_{\sfL_{\Phi}}(\sH) + 4\Rad_m^{\lambda}(\sH)\\
+ \frac{8 \max_{i} \abs*{\beta_i} B_{\Phi} }{m^2}
  + \bracket[\bigg]{2 B_{\lambda} + \frac{4 \max_{i} \abs*{\beta_i} B_{\Phi} }{m}}
  \sqrt{\tfrac{\log \frac{2 (\Delta \lambda) \, m}{\delta}}{2m}}}^{\frac{1}{2}}
\mspace {-10mu} + \frac{\ol_\beta}{m}.
\end{multline*}
}\fi
\end{restatable}
Then, for any $\delta >
0$, with probability at least $1 - \delta$, for all $\lambda \in
[\lambda_{\min}, \lambda_{\max}]$, we have
\begin{align*}
& \sE_{\ell^\lambda}(\h h_S) - \sE_{\ell^\lambda}^*(\sH)\\
& \leq O\paren[\Big]{
  \ov \Gamma
\paren[\Big]{\Rad_m^{\lambda}(\sH) +
  \textstyle
  \sqrt{\frac{\log ((\lambda_{\max} -  \lambda_{\min}) m/\delta)
    }{m}}
  + \sM_{\sfL_{\Phi}}(\sH) }
}.
\end{align*}
The proof is presented in
Appendix~\ref{app:generalization-uniform}.
We denote the right-hand of this bound by $\e_m$.  Building on the
ideas from Algorithm~\ref{alg:binary-search}, we introduce the
modified algorithm Algorithm~\ref{alg:binary-search-surrogate}.
Note that, with high probability, $\h \sE_{\ell^\lambda, S}(\h h_\lambda) >
\e_m$ implies $\sE_{\ell^\lambda}^*(\sH) > 0$ and, similarly, $\h
\sE_{\ell^\lambda, S}(\h h_\lambda) < -\e_m$ implies $\sE_{\ell^\lambda}^*(\sH) <
0$. Thus, this follows conditions used in the previous algorithm.  The
algorithm benefits from the following guarantee.

\begin{algorithm}[t]
  \caption{Generalized metrics optimization algorithm}
   \label{alg:binary-search-surrogate}
\begin{algorithmic}[1]
  \INPUT $\e$, $\e_m$.
  \STATE Initialize with $[a, b]
  = \bracket*{\lambda_{\min}, \lambda_{\max}}$
   \REPEAT
   \STATE  $\lambda \gets \frac{a + b}{2}$
   \STATE  $\h h_\lambda \gets \argmin_{h \in \sH} \h \sE_{\sfL_{\Phi}, S}(h)$
      \IF{($\h \sE_{\ell^\lambda, S}(\h h_\lambda) > \e_m$)}
         \STATE $[a, b] = [\lambda, b]$
      \ELSIF{($\h \sE_{\ell^\lambda, S}(\h h_\lambda) < -\e_m$)}
         \STATE $[a, b] = [a, \lambda]$
      \ELSE
         \STATE \RETURN $\h h_\lambda$
      \ENDIF
   \UNTIL{$\abs*{b - a} \leq \e$}
   \STATE \RETURN $\h h_\lambda$
\end{algorithmic}
\end{algorithm}

\begin{restatable}{theorem}{AlgoTwo}
\label{thm:algo2}
Let $\e = \frac{\e_m}{2 \ol_\beta}$. For any $\delta > 0$, with
probability at least $1 - \delta$,
Algorithm~\ref{alg:binary-search-surrogate} returns in $O
\paren*{\log_2\paren*{\frac{\lambda_{\max} - \lambda_{\min}}{\e}}}$
time a hypothesis $\h h_\lambda$ that admits the following guarantee:
\begin{equation*}
\cL(\h h_\lambda)
\leq  \cL^*(\sH) + \frac{4 \e_m}{\ul_\beta}.
\end{equation*}\\[-1.05cm]
\end{restatable}
The proof can be found in Appendix~\ref{app:algo2}. Thus, when $\e_m$
is small, that is, when the sample size is sufficiently large relative
to the complexity of $\sH$ and the minimizability gap is small, the
performance of $\h h_\lambda$ closely approaches the optimal
performance achievable with $\sH$.

In practice, the theoretical expression for $\e_m$ may not be
sufficiently tight due to constants or the minimizability gap, which
cannot be accurately approximated in non-realizable cases. In such
scenarios, $\lambda$ can be treated as a hyperparameter and tuned via
cross-validation, as outlined in Algorithm~\ref{alg:binary-search-cv}. 
\begin{algorithm}[h]
   \caption{Generalized metrics optimization algorithm with
     cross-validation}
   \label{alg:binary-search-cv}
\begin{algorithmic}[1]
\INPUT $\e$
   \STATE Initialize with $[a, b] = \bracket*{\lambda_{\min}, \lambda_{\max}}$, $\lambda^* = \lambda_{\max}$, $i = 0$
  \REPEAT
   \STATE  $\lambda \gets a + i \e$
    \STATE  $\h h_\lambda \gets \argmin_{h \in \sH} \h \sE_{\sfL_{\Phi}, S}(h)$
    \IF{($\h \cL_S(\h h_\lambda) < \lambda^*$)}
      \STATE $\h \lambda = \lambda$
      \STATE $\lambda^* \gets \h \cL_S(\h h_\lambda)$
      \ENDIF
    \STATE  $i \gets i + 1$
   \UNTIL{$a + i \e > b$}
   \STATE \RETURN $\h h_{\lambda}$
\end{algorithmic}
\end{algorithm}
The following result establishes
convergence and performance guarantees for this algorithm. Compared to
Algorithm~\ref{alg:binary-search-surrogate}, its computational
complexity is linear rather than logarithmic.

\begin{restatable}{theorem}{AlgoThree}
\label{thm:algo3}
For any $\delta > 0$, with probability at least $1 - \delta$, for $\e
\leq \frac{\e_m}{2 \ol_\beta}$, Algorithm~\ref{alg:binary-search-cv}
returns in $O \paren*{\frac{\lambda_{\max} - \lambda_{\min}}{\e}}$
time a hypothesis $\h h_\lambda$ that admits the following guarantee:
\[
\cL(\h h_\lambda) \leq \cL^*(\sH) + \frac{2\e_m}{\ul_\beta}.
\]\\[-1.05cm]
\end{restatable}
The proof is presented in Appendix~\ref{app:algo3}. We refer to Algorithms~\ref{alg:binary-search-surrogate} and \ref{alg:binary-search-cv} as \textbf{\METRO\ (\emph{Metric Optimization})}.  The effectiveness of \METRO\ compared to prior baselines is demonstrated by the experimental results reported in Section~\ref{sec:experiments}.

\ignore{Beyond the algorithms just presented, an iterative method can be
derived from Theorem~\ref{thm:lambda-star}.(see Appendix~\ref{app:iterative}).  While this approach
is computationally less efficient than binary search, it has potential
advantages, such as not requiring prior knowledge of the interval
containing $\lambda$.
}

Note that the quantities $\h h_\lambda$ and $\h \sE_{\ell^\lambda, S}(\h h_\lambda)$ in Algorithm~\ref{alg:binary-search-surrogate}, as well as  $\h h_\lambda$ and $\h \cL_S(\h h_\lambda)$ in Algorithm~\ref{alg:binary-search-cv},
 are approximated using data sampled from the same distribution, but they can be obtained from different samples. In practice, as done in \citep{KoyejoNatarajanRavikumarDhillon2014}, we can split the training data into two parts: $\h \lambda$
 is obtained from one part, and then used to train the hypothesis $\h h_{\h \lambda}$
 on the other.

Theorems~\ref{thm:algo2} and \ref{thm:algo3} remain valid as long as the data are sampled independently from the same distribution. However, in over-parameterized settings, the value of $\e_m$
 may be larger due to the high complexity of the model. This quantity
 becomes small only when the sample size is sufficiently large relative to the complexity of the hypothesis set. This limitation applies broadly to most generalization bounds for complex neural networks.

 \begin{figure}[t]
    \centering
    \vskip .05in
    \includegraphics[scale=0.4]{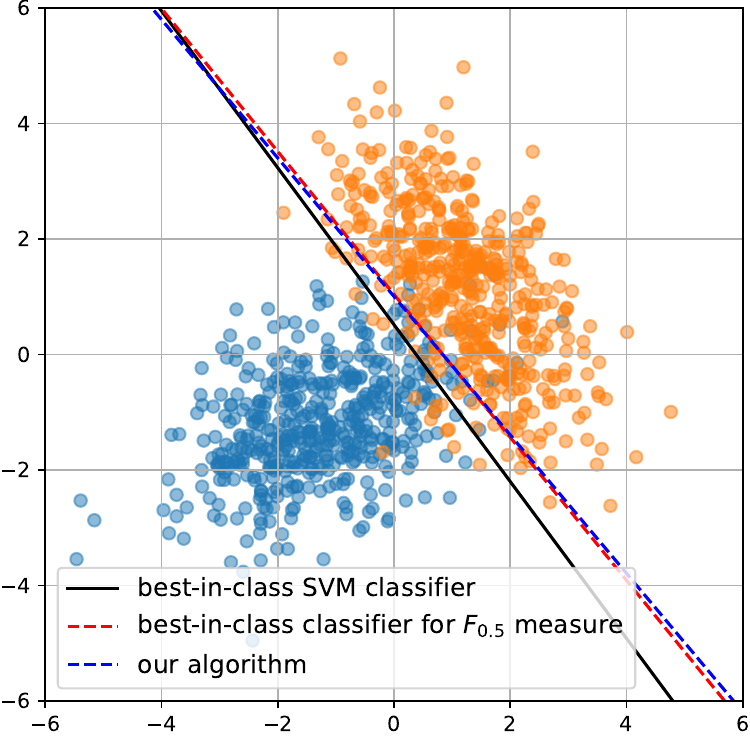}
    \caption{Comparison of the best-in-class SVM classifier, the
      optimal linear hypothesis for the $F_{0.5}$-measure, and the
      linear hypothesis returned by our algorithm.}
    \label{fig:example}
    \vskip -0.25in
\end{figure}

The current analysis of over-parameterized settings typically requires
alternative tools, particularly those that account for the
optimization algorithm (e.g., Stochastic Gradient Descent (SGD)
\citep{bottou2010large}) and its dynamics. Such analyses often apply
only to more restricted model families.

\textbf{Comparison of our algorithms with prior work.}  As discussed
in previous sections, our algorithms \METRO\ for optimizing general metrics
are supported by strong theoretical guarantees. These guarantees apply
to arbitrary hypothesis sets and provide finite-sample bounds. In
contrast, prior methods \citep{KoyejoNatarajanRavikumarDhillon2014}
rely solely on Bayes-consistency, which holds only for the class of
all measurable functions. This makes their analysis less relevant in
realistic settings where learning is restricted to specific function
classes, an issue explicitly left open in
\citep{KoyejoNatarajanRavikumarDhillon2014}. Furthermore, their
approach does not provide convergence rate guarantees, in contrast
with our finite-sample bounds.  In many modern applications,
particularly with complex neural networks, the minimizability gap is
often small or close to zero due to near-separable data. In such
cases, our learning guarantees become even more favorable.

Beyond theoretical advantages, a key limitation of prior methods is
their reliance on the structure of the Bayes-optimal solution. Since
the Bayes-optimal predictor for a given metric typically differs from
that of binary classification only by an offset, their approach first
trains a binary classifier and then selects an optimal threshold or
offset. However, this approach typically fails to find the best
predictor within a restricted hypothesis set.

Figure~\ref{fig:example} illustrates this issue with a simulated
example. The best-in-class linear hypothesis for $\cL$ with $\balpha =
-\paren*{\frac{5}{16}, \frac{5}{16}, \frac{5}{16}, \frac{5}{16}}$ and
$\bbeta = \paren*{0, \frac18, \frac12, \frac58}$ (corresponding to
the $F_{0.5}$-measure) is significantly different from the linear
hypothesis obtained by thresholding the best-in-class SVM
classifier. The two decision boundaries are not even parallel,
highlighting the fundamental inadequacy of thresholding-based
approaches such as those in
\citep{KoyejoNatarajanRavikumarDhillon2014}. In contrast, our
algorithm successfully finds a linear hypothesis that closely matches
the best-in-class solution for $\cL$.

Finally, note that when $\beta_1 = \beta_3 = 0$ for generalized
metrics, the independence of $\lambda^*$ from the hypothesis $h$ in
the target loss (Eq.~\eqref{eq:target-equiv-2}) significantly
simplifies our algorithm, as $\lambda^*$ no longer needs to be
estimated.

\section{Experiments}
\label{sec:experiments}

In this section, we present empirical results for our principled
algorithms for optimizing generalized metrics on the CIFAR-10
\citep{Krizhevsky09learningmultiple}, CIFAR-100
\citep{Krizhevsky09learningmultiple} and SVHN \citep{Netzer2011}
datasets.

\begin{table*}[t]
\caption{$F_{\beta}$ measure and JAC
  of three-hidden-layer neural network on CIFAR-10, CIFAR-100 and SVHN;
  mean $\pm$ standard deviation over five runs for ERM, Algorithm 1
  and Algorithm 2 in \citep{KoyejoNatarajanRavikumarDhillon2014},
  Algorithm in \citep{ParambathUsunierGrandvalet2014}, and \METRO\
  Algorithm.}
\label{tab:comparison}
\begin{center}
\resizebox{.9\textwidth}{!}{
    \begin{tabular}{@{\hspace{0pt}}lllll@{\hspace{0pt}}}
    Algorithm & Metric & CIFAR-10 & CIFAR-100  & SVHN \\
      \toprule
     ERM  & \multirow{5}{*}{$F_{1}$} & 0.9004 $\pm$ 0.0015   & 0.9265 $\pm$ 0.0067 &  0.9676 $\pm$ 0.0028 \\
     Algorithm 1 in \citep{KoyejoNatarajanRavikumarDhillon2014} & & 0.9040 $\pm$ 0.0134  & 0.9320 $\pm$ 0.0065 &  0.9679 $\pm$ 0.0035 \\
     Algorithm 2 in \citep{KoyejoNatarajanRavikumarDhillon2014} & & 0.9090 $\pm$ 0.0070  & 0.9317 $\pm$ 0.0114 & 0.9677 $\pm$ 0.0022 \\
     Algorithm in \citep{ParambathUsunierGrandvalet2014} & & 0.9185 $\pm$ 0.0029  & 0.9343 $\pm$ 0.0111 & 0.9682 $\pm$ 0.0021 \\
     \textbf{\METRO\   Algorithm} & & \textbf{0.9359 $\pm$ 0.0041}  & \textbf{0.9405 $\pm$ 0.0103} & \textbf{0.9713 $\pm$ 0.0029} \\
     \cmidrule{1-1} \cmidrule{2-5}
     ERM  & \multirow{5}{*}{$F_{0.5}$} & 0.9418 $\pm$ 0.0092   & 0.9338 $\pm$ 0.0216 &  0.9689 $\pm$ 0.0037 \\
     Algorithm 1 in \citep{KoyejoNatarajanRavikumarDhillon2014} & & 0.9476 $\pm$ 0.0032  & 0.9510 $\pm$ 0.0138 &  0.9715 $\pm$ 0.0035 \\
     Algorithm 2 in \citep{KoyejoNatarajanRavikumarDhillon2014} & & 0.9503 $\pm$ 0.0028  & 0.9435 $\pm$ 0.0140 & 0.9730 $\pm$ 0.0017 \\
     Algorithm in \citep{ParambathUsunierGrandvalet2014} & & 0.9507 $\pm$ 0.0036  & 0.9515 $\pm$ 0.0105 & 0.9746 $\pm$ 0.0011 \\
     \textbf{\METRO\   Algorithm} & & \textbf{0.9585 $\pm$ 0.0023}  & \textbf{0.9759 $\pm$ 0.0115} & \textbf{0.9807 $\pm$ 0.0015} \\
     \cmidrule{1-1} \cmidrule{2-5}
     ERM  & \multirow{5}{*}{$F_{1.5}$} & 0.9234 $\pm$ 0.0074   & 0.9279 $\pm$ 0.0160 &  0.9675 $\pm$ 0.0035 \\
     Algorithm 1 in \citep{KoyejoNatarajanRavikumarDhillon2014} & & 0.9305 $\pm$ 0.0029  & 0.9331 $\pm$ 0.0218 &  0.9688 $\pm$ 0.0025 \\
     Algorithm 2 in \citep{KoyejoNatarajanRavikumarDhillon2014} & & 0.9263 $\pm$ 0.0036  & 0.9340 $\pm$ 0.0119 & 0.9677 $\pm$ 0.0020 \\
     Algorithm in \citep{ParambathUsunierGrandvalet2014} & & 0.9312 $\pm$ 0.0044  & 0.9345 $\pm$ 0.0118 & 0.9702 $\pm$ 0.0023 \\
     \textbf{\METRO\   Algorithm} & & \textbf{0.9459 $\pm$ 0.0030}  & \textbf{0.9449 $\pm$ 0.0115} & \textbf{0.9771 $\pm$ 0.0018} \\
     \cmidrule{1-1} \cmidrule{2-5}
     ERM  & \multirow{5}{*}{JAC} & 0.4689 $\pm$ 0.0022   & 0.4693 $\pm$ 0.0063 &  0.4767 $\pm$ 0.0036 \\
     Algorithm 1 in \citep{KoyejoNatarajanRavikumarDhillon2014} & & 0.4821 $\pm$ 0.0024   & 0.4728 $\pm$ 0.0109 &  0.4814 $\pm$ 0.0031 \\
     Algorithm 2 in \citep{KoyejoNatarajanRavikumarDhillon2014} & & 0.5746 $\pm$ 0.0121  & 0.4753 $\pm$ 0.0119 & 0.4875 $\pm$ 0.0017 \\
     Algorithm in \citep{ParambathUsunierGrandvalet2014} & & 0.6428 $\pm$ 0.0130  & 0.4924 $\pm$ 0.0152 & 0.4934 $\pm$ 0.0019 \\
      \textbf{\METRO\ Algorithm} & & \textbf{0.6575 $\pm$ 0.0018} & \textbf{0.5017 $\pm$ 0.0046} & \textbf{0.4965 $\pm$ 0.0024} \\
    \end{tabular}
}
\end{center}
\vskip -0.25in
\end{table*}

Our experiments use a three-hidden-layer CNN with ReLU activations
\citep{lecun1995convolutional}. Standard data augmentations were
applied to CIFAR-10 and CIFAR-100, including 4-pixel padding followed
by $32 \times 32$ random cropping and random horizontal
flipping. Training was conducted using Stochastic Gradient Descent
(SGD) with Nesterov momentum \citep{nesterov1983method}. The initial
learning rate, batch size, and weight decay were set to $0.02$,
$1\mathord,024$, and $1 \times 10^{-4}$, respectively. A cosine decay
learning rate schedule \citep{loshchilov2022sgdr} was used over the
course of $100$ epochs. During training, we extract two classes from each
dataset to form a binary classification task.

We evaluated the models using their averaged generalized metric
$\cL$. In particular, we consider the $F_{\beta}$
measure \citep{ye2012optimizing}, where $\balpha = -\paren*{\frac{1 +
    \beta^2}{4}, \frac{1 + \beta^2}{4}, \frac{1 + \beta^2}{4}, \frac{1
    + \beta^2}{4}}$ and $\bbeta = \paren*{0, \frac{\beta^2}{2},
  \frac{1}{2}, \frac{\beta^2 + 1}{2}}$, and the Jaccard similarity
coefficient (JAC) \citep{sokolova2009systematic}, where $\balpha =
-\paren*{\frac{1}{4}, \frac{1}{4}, \frac{1}{4}, \frac{1}{4}}$ and
$\bbeta = \paren*{ \frac{1}{4}, \frac{1}{4}, -\frac{1}{4},
  \frac{3}{4}}$. 
The reported metric is averaged over five runs, with standard deviations included. We compared the \METRO\ algorithm using cross-validation (Algorithm~\ref{alg:binary-search-cv}) with four baselines: the standard
empirical risk minimization (ERM), the first algorithm (Algorithm 1)
and the second algorithm (Algorithm 2) from
\citep{KoyejoNatarajanRavikumarDhillon2014}, as well as the algorithm
from \citep{ParambathUsunierGrandvalet2014}, as detailed in
Section~\ref{sec:intro}. For \METRO\ algorithm, we used the surrogate loss
$\sfL_{\Phi}$ in \eqref{eq:sur} with the auxiliary function
$\Phi(t) = \log(1 + e^{-t})$, which corresponds to the logistic loss
function used in logistic regression. The logistic loss was also used
for ERM and Algorithm~1 in
\citep{KoyejoNatarajanRavikumarDhillon2014}, while Algorithm~2 in
\citep{KoyejoNatarajanRavikumarDhillon2014} and the algorithm from
\citep{ParambathUsunierGrandvalet2014} used the weighted logistic
loss. All the hyperparameters in these algorithms were selected
through cross-validation.

Table~\ref{tab:comparison} shows that our algorithm consistently
outperforms the four baselines across all datasets. Notably, the
algorithm in \citep{ParambathUsunierGrandvalet2014} consistently
outperforms Algorithms 1 and 2 in
\citep{KoyejoNatarajanRavikumarDhillon2014} by using two
hyperparameters: $\theta$ for weighting the logistic loss and
$\theta'$ for thresholding the classifier. In contrast, the relative
performance of Algorithms 1 and 2 in
\citep{KoyejoNatarajanRavikumarDhillon2014} varies across datasets,
although both outperform ERM.

Note that the per-epoch computational cost of our method is comparable
to that of Algorithm~2 in
\citep{KoyejoNatarajanRavikumarDhillon2014}. Both methods involve a
single hyperparameter, and for a fixed value of this parameter, the
computational cost is similar to training a standard binary classifier
using a standard surrogate loss.

In line with prior work \citep{KoyejoNatarajanRavikumarDhillon2014},
we used standard image classification datasets for our empirical
evaluation to demonstrate the effectiveness of our methods relative to
existing baselines. However, we acknowledge the importance of
evaluating our algorithms on more imbalanced datasets, which often
present greater challenges and are more representative of real-world
applications. We plan to include such experiments and comparisons in
future work.

One may wonder about the difficulty of directly optimizing a general
metric defined as the ratio of the expectations of two loss functions,
both linear in $\hh$, where $\hh(x) = \sign(h(x))$. While the metric
is quasi-concave in $\hh$, optimizing it with respect to $h$ is
NP-hard, even when the denominator is constant and $h$ is restricted
to a linear hypothesis set. In contrast, each surrogate loss
optimization problem we consider (framed as supervised learning) can
be solved in polynomial time over a convex hypothesis set, as the
surrogate loss functions we adopt are convex. Furthermore, directly
optimizing the empirical ratio of the numerator and denominator may
not yield a provably good approximation of the metric, since their
expectation does not align with the ratio of expectations.

\section{Conclusion}
\label{sec:conclusion}

We presented a series of theoretical, algorithmic, and empirical
results for optimizing generalized metrics in binary classification,
highlighting the significance of our algorithms supported by
$\sH$-consistency guarantees. Looking ahead, a natural
direction is extending our theory and algorithms to cover generalized
metrics in multi-class classification and multi-label learning,
broadening their applicability.

\section*{Acknowledgements}

We thank the anonymous reviewers for their valuable feedback and constructive suggestions.

\section*{Impact Statement}

This paper presents work whose goal is to advance the field of 
Machine Learning. There are many potential societal consequences 
of our work, none which we feel must be specifically highlighted here.

\bibliography{ghcb}
\bibliographystyle{icml2025}

\newpage
\appendix
\onecolumn

\renewcommand{\contentsname}{Contents of Appendix}
\tableofcontents
\addtocontents{toc}{\protect\setcounter{tocdepth}{3}} 
\clearpage

\section{Related work}
\label{app:related-work}

In binary classification, zero-one misclassification loss may not
always serve as an appropriate evaluation metric, particularly in
scenarios where more complex metrics are better suited to the
problem. For instance, in scenarios with significant class imbalance,
which frequently arise in applications such as fraud detection,
medical diagnosis, and text retrieval
\citep{lewis1995sequential,drummond2005severe,he2009learning,
  gu2009evaluation}, metrics like the \emph{$F_{\beta}$-measure}
\citep{lewis1995evaluating,jansche2005maximum,ye2012optimizing} and
the \emph{AM measure} \citep{menon2013statistical} are commonly
used. Similarly, \emph{weighted accuracy} is often used to address the
asymmetrical costs associated with different classes in real-world
applications. However, optimizing these generalized performance
metrics introduces both computational and statistical challenges, as
they do not conform to the standard single-loss expectations commonly
used for analyzing surrogate loss functions in machine learning
\citep{steinwart2007compare}.

Research on the design of algorithms for specific instances of such
metrics or the general family has largely focused on the
characterization of the Bayes-optimal classifier. For the standard
zero-one binary classification loss, it is known that the Bayes
classifier can be defined as $x \mapsto \sign(\eta(x) - \frac{1}{2})$,
where $\eta(x)$ is the probability of a positive label conditioned on
$x$.  A similar characterization applies to the weighted zero-one loss
\citep{scott2012calibrated}, where the threshold corresponds to a
weight different from $\frac{1}{2}$.  For imbalanced metrics,
\citet{ye2012optimizing} characterized the Bayes-classifier for the
$F_1$-measure, while \citet{menon2013statistical} characterized the
Bayes classifier for the AM measure as $x \mapsto \sign(\eta(x) -
\delta^*)$ for some $\delta^* \in (0, 1)$.  These results were later
generalized by \citet{KoyejoNatarajanRavikumarDhillon2014} to a
broader family of metrics that can be formulated as a ratio of two
linear functions of true positive (TP), false positive (FP), true
negative (TN), and false negative (FN) statistics. This family of
linear-fractional metrics includes the aforementioned weighted
accuracy, $F_{\beta}$-measures, the AM measure, as well as the
\emph{Jaccard similarity coefficient} \citep{sokolova2009systematic},
among others. Existing algorithms for optimizing these metrics heavily rely on the
structure of the Bayes-optimal solution
\citep{KoyejoNatarajanRavikumarDhillon2014,
  ParambathUsunierGrandvalet2014}. These methods generally adopt a
two-stage approach: first estimating the conditional probability
$\eta(x)$ and then searching for a suitable threshold $\delta^*$
specific to the metric.

In particular, \citet{KoyejoNatarajanRavikumarDhillon2014} study a general family of
performance metrics, including the $F_{\beta}$-measure, AM measure,
Jaccard similarity coefficient (JAC), and Weighted Accuracy
(WA). These metrics can be expressed as the ratio of two linear
combinations of four fundamental classification quantities. The
authors propose two relatively simple algorithms. Their first algorithm consists of training a standard binary
classifier such as logistic regression to return a real-valued
predictor $h$. Next, a threshold $\theta$ is chosen to maximize the
empirical $L$-measure for binary classifier $x \mapsto \sign(h(x) -
\theta)$, where $L$ represents the metric of interest. The authors do
not discuss how to find $\theta$ but presumably this can be done via
cross-validation based on a grid, or by binary search over all
possible values. Their second algorithm consists of training for each
fixed value of $\theta$ a cost-sensitive logistic regression (or other
margin-based algorithm) with weights $\theta$ and $(1 - \theta)$ and
return $h_\theta$.  Then, they find $\theta$ to maximize the empirical
$L$-measure for the binary classifier $x \mapsto
\sign(h_\theta(x))$. The authors provide a consistency-type guarantee
for these two-stage algorithms. \citet{ParambathUsunierGrandvalet2014} address the specific case of
the $F_{1}$-measure. Their algorithm coincides with the second
algorithm of \citet{KoyejoNatarajanRavikumarDhillon2014}.  However,
the authors also suggest returning $\sign(h_\theta(x) - \theta')$
where both $\theta$ and $\theta'$ are selected to maximize the
$L$-measure. The authors provide a stability-type guarantee for their
method, although the analysis lacks explicit details.

These algorithms naturally come with consistency guarantees \citep{Zhang2003,bartlett2006convexity,steinwart2007compare,KoyejoNatarajanRavikumarDhillon2014,MohriRostamizadehTalwalkar2018}. However, consistency is
an asymptotic property and does not provide explicit convergence rate
guarantees. Moreover, it applies only to the overly broad class of all
measurable functions, making it less relevant in practical scenarios
where learning is constrained to a restricted hypothesis class
\citep{long2013consistency,zhang2020bayes,awasthi2021calibration,awasthi2021finer,AwasthiMaoMohriZhong2023theoretically,awasthi2024dc,MaoMohriZhong2023rankingabs,MaoMohriZhong2023ranking,MaoMohriMohriZhong2023twostage,MaoMohriZhong2023characterization,MaoMohriZhong2023structured,zheng2023revisiting,MaoMohriZhong2024deferral,MaoMohriZhong2024predictor,MaoMohriZhong2024score,mao2024regression,mao2024h,mao2025enhanced,mao2024realizable,mao2024multi,MohriAndorChoiCollinsMaoZhong2023learning,cortes2024cardinality,CortesMaoMohriZhong2025balancing,cortes2025improved,MaoMohriZhong2025mastering,mao2025theory,montreuil2024learning,montreuil2025two,montreuil2025adversarial,montreuil2025ask,montreuil2025one,desalvo2025budgeted,mohri2025beyond,zhong2025fundamental}.

Other related work on generalized metrics includes studies on surrogate regret bounds \citep{reid2009surrogate,kotlowski2016surrogate}, extensions of the plug-in rule \citep{ye2012optimizing,dembczynski2013optimizing,narasimhan2014statistical,lipton2014thresholding,dembczynski2017consistency,yan2018binary,tavker2020consistent,berger2020threshold}, and structural loss optimization \citep{joachims2005support,kar2014online,yu2015learning,eban2017scalable,berman2018lovasz,bao2020calibrated}. Additionally, various optimization approaches have been explored, including online optimization \citep{kar2014online,zhang2018faster,kotlowski2024general,busa2015online} and constrained optimization \citep{narasimhan2019optimizing}. Further extensions address multi-class classification and multi-label learning \citep{dembczynski2013optimizing,narasimhan2015optimizing,ramaswamy2015consistent,narasimhan2015consistent,narasimhan2016optimizing,cheng2016efficient,natarajan2016optimal,sanyal2018optimizing,fathony2020ap,zhang2020convex,luo2021minimax,busa2022regret,schultheis2024consistent}, as well as generalized metrics under missing, corrupted, or noisy labels \citep{menon2015learning,natarajan2016regret,zhang2021learning,zhang2024multiclass}. Theoretical analyses and algorithms have also been developed for specific metrics, including the $F$-measure \citep{jansche2007maximum,jasinska2016extreme,pillai2017designing,bascol2019cost,jiang2020optimizing,berger2020threshold,zhang2020convex,dai2023rankseg}, fairness measures \citep{menon2018cost}, precision-recall \citep{flach2015precision}, and the balanced error rate (BER) \citep{zhao2013beyond}.

\section{Proof of Theorem~\ref{thm:equiv-non}}
\label{app:equiv-non}
\EquivNon*
\begin{proof}
Since we have $\E_{(x, y) \sim \sD} \bracket*{ \ell_{\bbeta}(h, x, y) } > 0$, the following equivalence holds for all $h \in \sH$ and $\eta \geq 0$:
\begin{align*}
\sE_{\ell^{\lambda^*\!\!}}(h) \leq \eta &\iff \E_{(x, y) \sim \sD} \bracket*{\ell_{\balpha}(h, x, y)} - \lambda^* \E_{(x, y) \sim \sD} \bracket*{ \ell_{\bbeta}(h, x, y)} \leq \eta \tag{def. of $\ell^{\lambda^*\!\!}$}\\
&\iff \frac{\E_{(x, y) \sim \sD} \bracket*{ \ell_{\balpha}(h, x, y) }}{\E_{(x, y) \sim \sD} \bracket*{ \ell_{\bbeta}(h, x, y) }} \leq \lambda^* + \frac{\eta}{\E_{(x, y) \sim \sD} \bracket*{ \ell_{\bbeta}(h, x, y) }} \tag{$\E_{(x, y) \sim \sD} \bracket*{ \ell_{\bbeta}(h, x, y) } > 0$}\\
&\iff \cL(h)  - \cL^*(\sH) \leq \frac{\eta}{\E_{(x, y) \sim \sD} \bracket*{ \ell_{\bbeta}(h, x, y) }} \tag{$\lambda^* = \cL^*(\sH)$}.
\end{align*}
This completes the proof.
\end{proof}

\section{Proof of Theorem~\ref{thm:bound}}
\label{app:bound}

We will use the following definitions. For any $x \in \sX$, we adopt
the definition $\eta(x) = \mathbb{P}(Y = +1 \!\mid\! X = x)$.  Then,
the \emph{conditional loss} of a hypothesis $h$ at point $x \in \sX$
for a loss function $\ell$ is defined as follows:
\[
\sC_{\ell}(h, x) = \eta(x) \ell(h, x, +1) + \paren*{1 - \eta(x)} \ell(h, x, -1).
\]
The \emph{best-in-class conditional loss} of a hypothesis set $\sH$ at
$x \in \sX$ is defined by $\sC^*_{\ell}(\sH, x) = \inf_{h \in \sH}
\sC_{\ell}(h, x)$ .
\begin{lemma}
\label{lemma:aux}
Assume that the following $\sH$-consistency bound holds for all $h \in \sH$ and all distributions:
\begin{equation*}
\sE_{\ell_{0-1}}(h) - \sE^*_{\ell_{0-1}}(\sH) + \sM_{\ell_{0-1}}(\sH) \leq \Gamma \paren*{\sE_{\Phi}(h) - \sE^*_{\Phi}(\sH) + \sM_{\Phi}(\sH)}.
\end{equation*}
Then, for any $\eta \in [0, 1]$ and $x \in \sX$, we have
\begin{align*}
& \ell_{0-1}(h, x, +1) \eta + \ell_{0-1}(h, x, -1) \paren*{1 - \eta} - \inf_{h \in \sH} \paren*{\ell_{0-1}(h, x, +1) \eta + \ell_{0-1}(h, x, -1) \paren*{1 - \eta} }\\
& \leq \Gamma\paren*{\Phi(h(x)) \eta + \Phi(-h(x)) \paren*{1 - \eta} - \inf_{h \in \sH} \paren*{\Phi(h(x)) \eta + \Phi(-h(x)) \paren*{1 - \eta} }}.
\end{align*}
\end{lemma}
\begin{proof}
For any $x \in \sX$, consider a distribution $\delta_{x}$ that concentrates on that point. Let $\eta = \mathbb{P}(Y = +1 \!\mid\! X = x)$. Then, by definition, $\sE_{\ell_{0-1}}(h) - \sE^*_{\ell_{0-1}}(\sH) + \sM_{\ell_{0-1}}(\sH)$ can be expressed as 
\begin{equation*}
\sE_{\ell_{0-1}}(h) - \sE^*_{\ell_{0-1}}(\sH) + \sM_{\ell_{0-1}}(\sH) = \ell_{0-1}(h, x, +1) \eta + \ell_{0-1}(h, x, -1) \paren*{1 - \eta} - \inf_{h \in \sH} \paren*{\ell_{0-1}(h, x, +1) \eta + \ell_{0-1}(h, x, -1) \paren*{1 - \eta} }.
\end{equation*}
Similarly, $\sE_{\Phi}(h) - \sE^*_{\Phi}(\sH) + \sM_{\Phi}(\sH)$ can be expressed as
\begin{equation*}
\Phi(h(x)) \eta + \Phi(-h(x)) \paren*{1 - \eta} - \inf_{h \in \sH} \paren*{\Phi(h(x)) \eta + \Phi(-h(x)) \paren*{1 - \eta} }.
\end{equation*}
Since the $\sH$-consistency bound holds by the assumption, we complete the proof.
\end{proof}

\Bound*
\begin{proof}
Let $\eta(x) =
\mathbb{P}(Y = 1 \!\mid\! X = x)$ be the conditional probability of $Y = 1$ given $X = x$.
By the definition, the conditional loss of the target loss can be expressed as follows: 
\begin{align*}
\sC_{\sfL}(h, x) &=  \eta(x) \sfL\paren*{\hh(x), +1} + (1 - \eta(x)) \sfL\paren*{\hh(x), -1}\\
& = 1_{\hh(x) = +1} \bracket*{\eta(x) \sfL(+1, +1) + (1 - \eta(x)) \sfL(+1, -1)} + 1_{\hh(x) = -1} \bracket*{\eta(x) \sfL(-1, +1) + (1 - \eta(x)) \sfL(-1, -1)}\\
& = \bracket*{\ell_{0-1}(h, x, +1) \eta'(x) + \ell_{0-1}(h, x, -1) \paren*{1 - \eta'(x)}} c(x),
\end{align*}
where $c(x) = \bracket*{\eta(x) \sfL(+1, +1) + (1 - \eta(x)) \sfL(+1, -1)} + \bracket*{\eta(x) \sfL(-1, +1) + (1 - \eta(x)) \sfL(-1, -1)} \in [0, 2\, \sfL_{\max}]$ and $\eta'(x) = \frac{\eta(x) \sfL(-1, +1) + (1 - \eta(x)) \sfL(-1, -1)}{c(x)} \in [0, 1]$.
Thus, the best-in-class conditional loss can be expressed as follows:
\begin{align*}
\sC^*_{\sfL}(\sH, x)  =  \inf_{h \in \sH} \bracket*{\ell_{0-1}(h, x, -1) \eta'(x) + \ell_{0-1}(h, x, -1) \paren*{1 - \eta'(x)}} c(x)
\end{align*}
By the definition,
\begin{align*}
& \Delta\sC_{\sfL, \sH}(h, x)\\
  & =   \sC_{\sfL}(h, x)- \sC^*_{\sfL}(\sH, x)\\
  & = \paren*{\bracket*{\ell_{0-1}(h, x, -1) \eta'(x) + \ell_{0-1}(h, x, -1) \paren*{1 - \eta'(x)}} - \inf_{h \in \sH} \bracket*{\ell_{0-1}(h, x, -1) \eta'(x) + \ell_{0-1}(h, x, -1) \paren*{1 - \eta'(x)}}} c(x).
\end{align*}
The conditional loss of the surrogate loss can be expressed as follows: 
\begin{align*}
\sC_{\sfL_{\Phi}}(h, x) & =  \eta(x) \paren*{\sfL(+1, +1) \Phi(-h(x)) + \sfL(-1, +1) \Phi(h(x))} + (1 - \eta(x)) \paren*{\sfL(+1, -1) \Phi(-h(x)) + \sfL(-1, -1) \Phi(h(x))}\\
& = \Phi(-h(x)) \bracket*{\eta(x) \sfL(+1, +1) + (1 - \eta(x)) \sfL(+1, -1)} +  \Phi(h(x)) \bracket*{\eta(x) \sfL(-1, +1) + (1 - \eta(x)) \sfL(-1, -1)}\\
& = \bracket*{\Phi(h(x)) \eta'(x) + \Phi(-h(x)) \paren*{1 - \eta'(x)}} c(x).
\end{align*}
Thus, the best-in-class conditional loss can be expressed as follows:
\begin{align*}
\sC^*_{\sfL_{\Phi}}(\sH, x)  =  \inf_{h \in \sH} \bracket*{\Phi(h(x)) \eta'(x) + \Phi(-h(x)) \paren*{1 - \eta'(x)}} c(x).
\end{align*}
By the definition,
\begin{align*}
& \Delta\sC_{\sfL_{\Phi}, \sH}(h, x)\\
  & =   \sC_{\sfL_{\Phi}}(h, x)- \sC^*_{\sfL_{\Phi}}(\sH, x)\\
  & = \paren*{\bracket*{\Phi(h(x)) \eta'(x) + \Phi(-h(x)) \paren*{1 - \eta'(x)}} - \inf_{h \in \sH} \bracket*{\Phi(h(x)) \eta'(x) + \Phi(-h(x)) \paren*{1 - \eta'(x)}}} c(x).
\end{align*}
By Lemma~\ref{lemma:aux}, we have
\begin{align*}
& \Delta\sC_{\sfL_{\Phi}, \sH}(h, x)\\
& \paren*{\bracket*{\Phi(h(x)) \eta'(x) + \Phi(-h(x)) \paren*{1 - \eta'(x)}} - \inf_{h \in \sH} \bracket*{\Phi(h(x)) \eta'(x) + \Phi(-h(x)) \paren*{1 - \eta'(x)}}} c(x)\\
& \geq \Gamma^{-1} \paren*{\bracket*{\ell_{0-1}(h, x, -1) \eta'(x) + \ell_{0-1}(h, x, -1) \paren*{1 - \eta'(x)}} - \inf_{h \in \sH} \bracket*{\ell_{0-1}(h, x, -1) \eta'(x) + \ell_{0-1}(h, x, -1) \paren*{1 - \eta'(x)}}} c(x)\\
& = \frac{1}{\beta^{\frac1{\alpha}}} \paren*{\bracket*{\ell_{0-1}(h, x, -1) \eta'(x) + \ell_{0-1}(h, x, -1) \paren*{1 - \eta'(x)}} - \inf_{h \in \sH} \bracket*{\ell_{0-1}(h, x, -1) \eta'(x) + \ell_{0-1}(h, x, -1) \paren*{1 - \eta'(x)}}}^{\frac1{\alpha}} c(x)\\
& = \frac{1}{\beta^{\frac1{\alpha}}} \paren[\Big]{\paren[\big]{\bracket*{\ell_{0-1}(h, x, -1) \eta'(x) + \ell_{0-1}(h, x, -1) \paren*{1 - \eta'(x)}}\\
&\qquad - \inf_{h \in \sH} \bracket*{\ell_{0-1}(h, x, -1) \eta'(x) + \ell_{0-1}(h, x, -1) \paren*{1 - \eta'(x)}}} c(x)}^{\frac1{\alpha}} \paren*{c(x)}^{1 - \frac{1}{\alpha}}\\
& \geq \frac{1}{\beta^{\frac1{\alpha}}} \paren*{\Delta\sC_{\sfL, \sH}(h, x)}^{\frac1{\alpha}} \paren*{2 \, \sfL_{\max}}^{1 - \frac{1}{\alpha}}.
\end{align*}
By taking the expectation on both sides and applying Jensen's inequality, we obtain:
\begin{align*}
\sE_{\sfL}(h) - \sE_{\sfL}^*(\sH) + \sM_{\sfL}(\sH) \leq \ov \Gamma \paren*{\sE_{\sfL_{\Phi}}(h) - \sE_{\sfL_{\Phi}}^*(\sH) + \sM_{\sfL_{\Phi}}(\sH)},
\end{align*}
where $\ov \Gamma(t) = \beta  \paren*{2 \, \sfL_{\max}}^{1 - \alpha} t^{\alpha}$.
\end{proof}

\section{Proof of Theorem~\ref{thm:lambda-star}}
\label{app:lambda-star}

\LambdaStar*
\begin{proof}
  For any $h \in \sH$, define $f(h)$ and $g(h)$ to simplify notation:
  \begin{align*}
    f(h)
    = \frac{\E_{(x, y) \sim \sD} \bracket*{\ell_{\balpha}(h, x, y)}}{\E_{(x, y) \sim \sD} \bracket*{\ell_{\bbeta}(h, x, y)}},
\quad g(h)
= \E_{(x, y) \sim \sD} \bracket*{\ell_{\bbeta}(h, x, y)}.
\end{align*}
  By assumption, we have $g(h) > 0$ for all $h \in \sH$ and $\lambda^*
  = \inf_{h \in \sH} f(h)$ and $\sE_{\ell^{\lambda^*\!\!}}^*(\sH) = \inf_{h \in \sH} \curl*{(f(h) - \lambda^*)
    g(h)}$. Since $g$ is upper-bounded by $\ol_\beta$ and $(f(h) -
  \lambda^*) \geq 0$, it follows that
  \[
  \sE_{\ell^{\lambda^*\!\!}}^*(\sH)
  \leq \inf_{h \in \sH} \curl*{f(h) - \lambda^*} \, \ol_\beta = 0.
  \]
By definition of $\sE_{\ell^{\lambda^*\!\!}}^*(\sH)$ as
an infimum, for any $\eta > 0$, there exists $h_\eta \in \sH$ such
that
\[
\sE_{\ell^{\lambda^*\!\!}}^*(\sH) + \eta
> (f(h_\eta) - \lambda^*) g(h_\eta) \geq 0.
\]
Since $\sE_{\ell^{\lambda^*\!\!}}^*(\sH) + \eta > 0$ for
all $\eta > 0$, it follows that $\sE_{\ell^{\lambda^*\!\!}}^*(\sH) \geq 0$.  Combining the two inequalities yields
$\sE_{\ell^{\lambda^*\!\!}}^*(\sH) = 0$. This
establishes the first equality.

Next, assume $\lambda^* - \lambda > 0$.  By the definition of
$\lambda^*$ as an infimum, we have $f(h) - \lambda > 0$ for all $h \in
\sH$. This implies $\sE_{\ell^\lambda}^*(\sH) =
\inf_{h \in \sH} \curl*{(f(h) - \lambda) g(h)} \geq \ul_\beta \inf_{h
  \in \sH} \curl*{(f(h) - \lambda)} = \ul_\beta (\lambda^* - \lambda) >
0$, thus $\sE_{\ell^\lambda}^*(\sH) > 0$, which
proves one direction of the statement.

Now, assume $\lambda^* - \lambda < 0$.  By the definition of
$\lambda^*$ as an infimum, for any $\eta > 0$, there exists $h_\eta
\in \sH$ such that $f(h_\eta) < \lambda^* + \eta$.  Choose $\eta <
(\lambda - \lambda^*)$.  This implies $\sE_{\ell^{\lambda}}^*(\sH)
\leq (f(h_\eta) - \lambda) g(h_\eta) \leq (\lambda^* + \eta - \lambda)
g(h_\eta)$. Since $(\lambda^* + \eta - \lambda) < 0$ and $g(h_\eta) >
0$, it follows that $\sE_{\ell^\lambda}^*(\sH) < 0$. This proves the
other direction.
\end{proof}

\section{Proof of Theorem~\ref{thm:rate}}
\label{app:rate}

\begin{lemma}
\label{lemma:approx}
Fix $\e > 0$ and assume $\abs*{\lambda - \lambda^*} \leq \e$.
Then, the following inequality holds:
\[
\sE_{\ell^\lambda}^*(\sH) \leq  \e \ol_\beta.
\]
\end{lemma}
\begin{proof}
By definition of $\sE_{\ell^\lambda}(h)$, the following
holds for any $h \in \sH$:
\begin{align*}
\sE_{\ell^\lambda}(h) 
& = \sE_{\ell^{\lambda^*\!\!}}(h)
+ (\lambda^* - \lambda) \E_{(x, y) \sim \sD}[\ell_\beta(h, x, y)]
\leq \sE_{\ell^{\lambda^*\!\!}}(h) + \e \ol_\beta.
\end{align*}
Thus, by definition of $\sE_{\ell^\lambda}^*(\sH)$
as an infimum, for any $h \in \sH$, we have
$\sE_{\ell^\lambda}^*(\sH)\leq
\sE_{\ell^{\lambda^*\!\!}}(h) + \e \ol_\beta$. Taking the
infimum of the right-hand side over $\sH$ yields
$\sE_{\ell^\lambda}^*(\sH) \leq
\sE_{\ell^{\lambda^*\!\!}}^*(\sH) + \e \ol_\beta$.  By
Theorem~\ref{thm:lambda-star}, we have
$\sE_{\ell^{\lambda^*\!\!}}^*(\sH) = 0$. This completes
the proof.
\end{proof}

\Rate*

\begin{proof}
  By definition of the binary search, we have $|\lambda - \lambda^*| \leq \e$.
  Thus, the inequality holds by Lemma~\ref{lemma:approx}. The time complexity
  follows straightforwardly the property of the binary search.
\end{proof}

\section{Proof of Theorem~\ref{thm:generalization}}
\label{app:generalization}

\Generalization*
\begin{proof}
  By the standard Rademacher complexity bounds \citep{MohriRostamizadehTalwalkar2018}, the following holds
  with probability at least $1 - \delta$ for all $h \in \sH$:
\[
\abs*{\sE_{\sfL_{\Phi}}(h) - \h \sE_{\sfL_{\Phi}, S}(h)}
\leq 2 \Rad_m^\lambda(\sH) +
B_\lambda \sqrt{\tfrac{\log (2/\delta)}{2m}}.
\]
Fix $\e > 0$. By the definition of the infimum, there exists $h^* \in
\sH$ such that $\sE_{\sfL_{\Phi}}(h^*) \leq
\sE_{\sfL_{\Phi}}^*(\sH) + \e$. By definition of
$\h h_\lambda$, we have
\begin{align*}
  & \sE_{\sfL_{\Phi}}(\h h_\lambda) - \sE_{\sfL_{\Phi}}^*(\sH)\\
  & = \sE_{\sfL_{\Phi}}(\h h_\lambda) - \h \sE_{\sfL_{\Phi}, S}(\h h_\lambda) + \h \sE_{\sfL_{\Phi}, S}(\h h_\lambda) - \sE_{\sfL_{\Phi}}^*(\sH)\\
  & \leq \sE_{\sfL_{\Phi}}(\h h_\lambda) - \h \sE_{\sfL_{\Phi}, S}(\h h_\lambda) + \h \sE_{\sfL_{\Phi}, S}(h^*) - \sE_{\sfL_{\Phi}}^*(\sH)\\
  & \leq \sE_{\sfL_{\Phi}}(\h h_\lambda) - \h \sE_{\sfL_{\Phi}, S}(\h h_\lambda) + \h \sE_{\sfL_{\Phi}, S}(h^*) - \sE_{\sfL_{\Phi}}^*(h^*) + \e\\
  & \leq
  2 \bracket*{2 \Rad_m^\lambda(\sH) +
B_\lambda \sqrt{\tfrac{\log (2/\delta)}{2m}}} + \e.
\end{align*}
Since the inequality holds for all $\e > 0$, it implies:
\[
\sE_{\sfL_{\Phi}}(\h h_\lambda) - \sE_{\sfL_{\Phi}}^*(\sH)
\leq 
4 \Rad_m^\lambda(\sH) +
2 B_\lambda \sqrt{\tfrac{\log (2/\delta)}{2m}}.
\]
Substituting this inequality into the $\sH$-consistency bound in the assumption completes the proof.
\end{proof}

\section{Proof of Theorem~\ref{thm:generalization-uniform}}
\label{app:generalization-uniform}

Let $B_{\Phi}$ be
an upper bound for the function $\Phi$. This boundedness holds in practice when we consider a bounded input space.
\begin{lemma}
\label{lemma:approx-Rad}
Fix $\e > 0$ and assume $\abs*{\lambda_1 - \lambda_2} \leq \e$.
Then, the following inequalities hold:
\[
\abs*{\Rad_m^{\lambda_1}(\sH) - \Rad_m^{\lambda_2}(\sH)} \leq \frac{4 \max_{i} \abs*{\beta_i} B_{\Phi} \e}{m} , \quad \abs*{B_{\lambda_1} - B_{\lambda_2}} \leq 4 \max_{i} \abs*{\beta_i} B_{\Phi} \e.
\]
\end{lemma}
\begin{proof}
Let $\sfL_{\Phi}^{\lambda}$ be the surrogate loss \eqref{eq:sur} with the parameter $\lambda$. By the definition of the cost function \eqref{eq:target-equiv-2}, the following
holds for any $(h, x, y) \in \sH \times \sX \times \sY$:
\begin{align*}
\abs*{\sfL_{\Phi}^{\lambda_1}(h, x, y) - \sfL_{\Phi}^{\lambda_2}(h, x, y)} \leq 4 \max_{i} \abs*{\beta_i} B_{\Phi} \abs*{\lambda_1 - \lambda_2} .
\end{align*}
Thus, by definition of Rademacher complexity and using $B_{\lambda} = \sup_{h, x, y }\sfL_{\Phi}(h, x, y)$, we have
\[
\abs*{\Rad_m^{\lambda_1}(\sH) - \Rad_m^{\lambda_2}(\sH)} \leq \frac{4 \max_{i} \abs*{\beta_i} B_{\Phi} \e}{m}, \quad \abs*{B_{\lambda_1} - B_{\lambda_2}} \leq 4 \max_{i} \abs*{\beta_i} B_{\Phi} \e.
\]
This completes
the proof.
\end{proof}

\GeneralizationUniform*
\begin{proof}
To derive the uniform bound, we cover
the interval $[\lambda_{\min},
  \lambda_{\max}]$ using sub-intervals of size $1/m$.
Consider sequences $\paren*{\lambda_{k}}_{k}$, where $\lambda_k = \lambda_{\min} + \frac{k - 1}{m}$, $1 \leq k \leq m \paren*{\lambda_{\max} - \lambda_{\min}}$. By Theorem~\ref{thm:generalization} and the standard uniform bounds \citep{MohriRostamizadehTalwalkar2018}, for any $\delta > 0$, with probability at least $1 - \delta$, the following bound holds for all $k$:
\begin{equation*}
\sE_{\lambda_k}(\h h_S) - \sE_{\lambda_k}^*(\sH)
\leq \Gamma
\paren[\bigg]{\sM_{\sfL_{\Phi}}(\sH) + 4 \Rad_m^{\lambda_k}(\sH)
  + 2 B_{\lambda_k} \sqrt{\tfrac{\log \frac{2 (\lambda_{\max} - \lambda_{\min}) \, m}{\delta}}{2m}}}.
\end{equation*}
Then, for any $\lambda \in [\lambda_{\min},
  \lambda_{\max}]$, there exists $k \geq 1$ such that $\abs*{\lambda - \lambda_k} \leq \frac{1}{2m}$. By Lemma~\ref{lemma:approx} and Lemma~\ref{lemma:approx-Rad}, we obtain
\begin{equation*}
\sE_{\ell^\lambda}(\h h_S) - \sE_{\ell^\lambda}^*(\sH)
\leq \ov \Gamma
\paren[\bigg]{\sM_{\sfL_{\Phi}}(\sH) + 4\Rad_m^{\lambda}(\sH) + \frac{8 \max_{i} \abs*{\beta_i} B_{\Phi} }{m^2}
  + \paren*{2 B_{\lambda} + \frac{4 \max_{i} \abs*{\beta_i} B_{\Phi} }{m}}\sqrt{\tfrac{\log \frac{2 (\lambda_{\max} - \lambda_{\min}) \, m}{\delta}}{2m}}} + \frac{\ol_\beta}{m}.
\end{equation*}  
This completes the proof of the first bound. The second bound follows
directly by substituting $\ov \Gamma (t) = 2 \paren*{
  \sfL_{\max}}^{\frac{1}{2}} t^{\frac{1}{2}}$ into the first bound.
\end{proof}

\section{Proof of Theorem~\ref{thm:algo2}}
\label{app:algo2}

\AlgoTwo*
\begin{proof}
Suppose that $\h h_\lambda$ is the solution returned at step 10 of
Algorithm~\ref{alg:binary-search-surrogate}.  Then, by definition of
the algorithm, we must have $|\h \sE_{\ell^{\lambda}}(\h h_\lambda)|
\leq \e_m$, which, by the standard generalization bound
\citep{MohriRostamizadehTalwalkar2018}, implies that (with high
probability)
\[
\sE_{\ell^\lambda}(\h h_\lambda)
\leq 2 \e_m
\]
Expanding the definition of the surrogate risk, this gives
\[
\E[\ell_\alpha(h, x, y)] - \lambda \E[\ell_\beta(h,
  x, y)] \leq 2 \e_m.
\]
Dividing through by $\E[\ell_\beta(h, x, y)]$ yields
\begin{equation}
\label{eq:intermediate}
\cL(\h h_\lambda)
\leq \lambda + \frac{2 \e_m}{\E[\ell_\beta(h,
    x, y)]} \leq \lambda + \frac{2 \e_m}{\ul_\beta}.
\end{equation}
Next, observe that for all $h \in \sH$, we have
\[
\sE_{\ell^\lambda}(h) - \sE_{\ell^{\lambda^*\!\!}}(h)
= (\lambda - \lambda^*) \E[\ell_\beta(h, x, y)].
\]
Rearranging gives
\[
(\lambda - \lambda^*)
= \frac{\sE_{\ell^\lambda}(h)
  - \sE_{\ell^{\lambda^*\!\!}}(h)}{\E[\ell_\beta(h, x, y)]}.
\]
Since $\sE_{\ell^{\lambda^*\!\!}}(\sH) = 0$, by definition of the infimum, it follows that $\sE_{\ell^{\lambda^*\!\!}}(h) \geq 0$. 
In view of that, we can write:
\[
(\lambda - \lambda^*)
\leq \frac{\sE_{\ell^\lambda}(h)}{\E[\ell_\beta(h, x, y)]}.
\]
Applying this inequality with $h = \h h_\lambda$ and substituting into \eqref{eq:intermediate}, we obtain  
\[
\cL(\h h_\lambda) \leq \lambda^*
+ \frac{2 \e_m}{\ul_\beta}
+ \frac{\sE_{\ell^\lambda}(\h h_\lambda)}{\E[\ell_\beta(\h h_\lambda, x, y)]}
  \leq \lambda^* + \frac{4 \e_m}{\ul_\beta}.
\]
This ends the analysis of that case.

Now, consider the case where the algorithm terminates with $|b - a|
\leq \e_m$. In this case, we have (with high probability) $|\lambda -
\lambda^*| \leq \e$. By Lemma~\ref{lemma:approx}, this implies
\[
\sE^*_{\ell^{\lambda}}(\sH) \leq \e \ol_\beta.
\]
Combining this with the estimation bound for $\ell^{\lambda}$, we have (with high probability)
\[
\sE_{\ell^\lambda}(\h h_\lambda)
\leq \sE^*_{\ell^{\lambda}}(\sH) + \e_m \leq \e \ol_\beta + \e_m.
\]
Thus,
\[
\E[\ell^{\lambda}_{\balpha}(\h h_\lambda, x, y)]
- \lambda \E[\ell^{\lambda}_{\beta}(\h h_\lambda, x, y)]
\leq \e \ol_\beta + \e_m.
\]
  Dividing by $\E[\ell^{\lambda}_{\beta}(\h h_\lambda, x, y)]$ both sides
  yields
  \[
  \cL_{\alpha, \beta}(\h h_\lambda) \leq \lambda + \frac{\e \ol_\beta
    + \e_m}{\ul_\beta} \leq \lambda^* + \e + \frac{\e \ol_\beta
    + \e_m}{\ul_\beta}.
  \]
  Choosing $\e = \e_m/(2 \ol_\beta)$ yields
  \[
  \cL_{\alpha, \beta}(\h h_\lambda) \leq \lambda + \frac{\e \ol_\beta
    + \e_m}{\ul_\beta} \leq \lambda^* + \frac{\e_m}{2 \ol_\beta} + \frac{3
    \e_m}{2\ul_\beta} \leq \lambda^* + \frac{2\e_m}{\ul_\beta}.
  \]
  This completes the proof.
\end{proof}

\section{Proof
  of Theorem~\ref{thm:algo3}}
\label{app:algo3}

\AlgoThree*
\begin{proof}
Since the algorithm terminates with $a + i \e > b$, we have $|\lambda -
\lambda^*| \leq \e$. By Lemma~\ref{lemma:approx}, this implies
\[
\sE^*_{\ell^{\lambda}}(\sH) \leq \e \ol_\beta.
\]
Combining this with the estimation bound for $\ell^{\lambda}$, we have (with high probability)
\[
\sE_{\ell^\lambda}(\h h_\lambda)
\leq \sE^*_{\ell^{\lambda}}(\sH) + \e_m \leq \e \ol_\beta + \e_m.
\]
Thus,
\[
\E[\ell^{\lambda}_{\balpha}(\h h_\lambda, x, y)]
- \lambda \E[\ell^{\lambda}_{\beta}(\h h_\lambda, x, y)]
\leq \e \ol_\beta + \e_m.
\]
  Dividing by $\E[\ell^{\lambda}_{\beta}(\h h_\lambda, x, y)]$ both sides
  yields
  \[
  \cL_{\alpha, \beta}(\h h_\lambda) \leq \lambda + \frac{\e \ol_\beta
    + \e_m}{\ul_\beta} \leq \lambda^* + \e + \frac{\e \ol_\beta
    + \e_m}{\ul_\beta}.
  \]
  Choosing $\e \leq \e_m/(2 \ol_\beta)$ yields
  \[
  \cL_{\alpha, \beta}(\h h_\lambda) \leq \lambda + \frac{\e \ol_\beta
    + \e_m}{\ul_\beta} \leq \lambda^* + \frac{\e_m}{2 \ol_\beta} + \frac{3
    \e_m}{2\ul_\beta} \leq \lambda^* + \frac{2\e_m}{\ul_\beta}.
  \]
  This completes the proof.
\end{proof}

\end{document}